\documentclass[final]{cvpr}

\usepackage{amsmath,amsfonts,bm}

\def\eqref#1{equation~\ref{#1}}

\def\1{\bm{1}}

\def\rmI{{\mathbf{I}}}

\def\vp{{\bm{p}}}

\def\vu{{\bm{u}}}
\def\vv{{\bm{v}}}

\def\vx{{\bm{x}}}

\def\mB{{\bm{B}}}

\def\mM{{\bm{M}}}

\DeclareMathAlphabet{\mathsfit}{\encodingdefault}{\sfdefault}{m}{sl}
\SetMathAlphabet{\mathsfit}{bold}{\encodingdefault}{\sfdefault}{bx}{n}

\newcommand{\E}{\mathbb{E}}

\newcommand{\R}{\mathbb{R}}

\usepackage{times}
\usepackage{epsfig}
\usepackage{graphicx}
\usepackage{amsmath, bm}
\usepackage{amssymb}
\usepackage{caption}
\usepackage{subcaption}
\usepackage{booktabs}
\usepackage{multirow}
\usepackage{siunitx}
\usepackage{url}
\newcommand\norm[1]{\left\lVert#1\right\rVert}

\usepackage[pagebackref=true,breaklinks=true,colorlinks,bookmarks=false]{hyperref}

\def\cvprPaperID{5721} 
\def\confYear{CVPR 2021}

\begin{document}
\pagestyle{empty}
\title{\ Learning Neural Representation of Camera Pose with Matrix Representation of Pose Shift via View Synthesis}\pagestyle{empty}

\author{Yaxuan Zhu, Ruiqi Gao, Siyuan Huang, Song-Chun Zhu, and Ying Nian Wu\\
Department of Statistics, University of California, Los Angeles (UCLA)\\
{\tt\small \{yaxuanzhu, ruiqigao, huangsiyuan\}@ucla.edu, \{sczhu, ywu\}@stat.ucla.edu} \\
}\pagestyle{empty}

\clearpage\maketitle
\pagestyle{empty}

\begin{abstract}
   How to effectively represent camera pose is an essential problem in 3D computer vision, especially in tasks such as  camera pose regression and novel view synthesis. Traditionally, 3D position of the camera is represented by Cartesian coordinate and the orientation is represented by Euler angle or quaternions. These representations are manually designed, which may not be the most effective representation for downstream tasks. In this work, we propose an approach to learn neural representations of camera poses and 3D scenes, coupled with neural representations of local camera movements. Specifically, the camera pose and 3D scene are represented as vectors and the local camera movement is represented as a matrix operating on the vector of the camera pose. We demonstrate that the camera movement can further be parametrized by a matrix Lie algebra that underlies a rotation system in the neural space. The vector representations are then concatenated and generate the posed 2D image through a decoder network. The model is learned from only posed 2D images and corresponding camera poses, without access to depths or shapes. We conduct extensive experiments on synthetic and real datasets. The results show that compared with other camera pose representations, our learned representation is more robust to noise in novel view synthesis and more effective in camera pose regression.
\end{abstract}

\section{Introduction}
With the advance of deep neural network (DNN), there has been a series of successful works that employ DNN in camera pose estimation \cite{kendall2015posenet, kendall2017geometric, brahmbhatt2018geometry,saha2018improved,balntas2018relocnet, laskar2017camera} or object pose estimation \cite{do2018deep}. In contrast, novel view synthesis is in the opposite direction that maps the camera pose and 3D scene representation back to the posed 2D image under certain view~\cite{eslami2018neural, sitzmann2019scene}. A fundamental problem in both lines of work is to find effective representations of the camera pose~\cite{zhou2019continuity}. Existing methods include representing the agent's position in 3D Cartesian coordinate, and the 3D 
orientation can be represented by Euler angle, axis-angle, $SO$(3) rotation matrices, quaternions or log quaternions. These representations are mainly defined in manually designed coordinates where each dimension has highly abstract semantics, which could be suboptimal when involved in the optimization with deep neural networks. It is desirable to have learning-based representations for camera poses.

Recently, \cite{gao2018learning} proposes a representational model of grid cells in the entorhinal cortex of mammalian brains. Grid cells have been found participating in mental self-navigation and they fire at strikingly regular hexagon grids of positions when the agent moves within an open field. The representational model in~\cite{gao2018learning} consists of a vector representation of agent's self-position, coupled with a matrix representation of agent's self-motion. When the agent undergoes a certain self-motion in the 2D space, the vector of self-position is rotated by the matrix of self-motion on a 2D sub-manifold in the mental space. Such a model achieves self-navigation and learns hexagon grid patterns of grid cells, which has the promise to be biologically plausible. 

\begin{figure}[t!]
    \centering
    \includegraphics[width=0.4\textwidth]{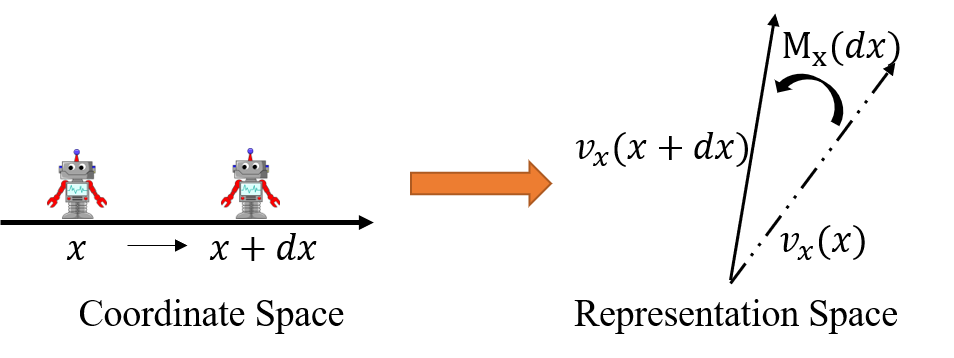}
    \caption{\small Illustration of our proposed pose representation. Take axis $x$ as an example. The agent's position on axis $x$ is mapped to a high dimensional vector and the agent's movement along axis $x$ is modeled as a rotation of the vector.}
    \label{fig:illustrate idea}
\end{figure}

Inspired by~\cite{gao2018learning,gao2020path}, we propose an approach towards learning neural representation of camera pose, coupled with representation of local camera movement. Specifically, given 2D posed images of a 3D scene and their corresponding camera poses, we assume a shared vector representation for the underlying 3D scene and a distinct vector representation for the camera pose of each 2D image. When the camera has a local displacement, the vector of 3D scene remains unchanged while the vector of camera pose is rotated by the matrix representation of camera movement (Figure \ref{fig:illustrate idea}). We further parametrize the matrix representation by matrix Lie group and the corresponding matrix Lie algebra. The vector representations of camera poses and matrix presentations of camera movements can be shared across multiple scenes, so that they can be learned from multiple scenes to boost performance. The vectors of 3D scene and camera pose are concatenated together to generate the 2D image through a decoder network (Figure \ref{fig:illustration frame-work}). The model is learned with only posed 2D images and camera poses, without extra knowledge such as depths or shapes. We perform various experiments on synthetic and real datasets in the context of novel view synthesis and camera pose regression. 

The contributions of our work include:

\begin{enumerate}
    \item We propose a method for learning neural camera pose representation coupled with neural camera movement representation.
    \item We associate this representational model with the agent's visual input through a generative model. 
    \item We demonstrate that the learned neural representation is  effective as the target representation in camera pose regression. 
\end{enumerate}

\section{Related work}
\subsection{Representing camera orientation and position}
The simplest way to represent orientation is by Euler angle. However, as \cite{kendall2015posenet, kendall2017geometric} point out, Euler angle wraps around at $2\pi$ and is not injective to 3D rotation, and thus can be difficult to learn. \cite{balntas2018relocnet} uses $SO$(3) rotation matrices to represent the relative orientation rotation between a pair of images. $SO$(3) rotation matrices are an over-parameterized representation of rotation which has the property of orthonormality. However, it is in general difficult to enforce the orthonormality constraint when learning a $SO$(3) representation through back-propagation. \cite{ummenhofer2017demon, mahendran20173d} use axis-angle representation, which represents 3D orientation by the direction of axis of rotation as well as the magnitude of the rotation. Similar to Eluer angles, this representation also has the problem of repetition around the $2\pi$ radians. PoseNet and its variants \cite{kendall2015posenet, kendall2017geometric} propose to use quaternions. Quarernions, or more specfically, quaternions with unit length, are a 4-D continuous and smooth representation of rotation. MapNet \cite{brahmbhatt2018geometry} further proposes to use log quaternions to avoid over-parametrization. These quaternion-based methods achieve state-of-the-art results in the area of absolute camera pose regression. \cite{zhou2019continuity} argues that these representations are not continuous and  proposes another 5D or 6D representation for orientation. All these representations are manually designed and pre-defined. \cite{levinson2020analysis} introduces SVD orthogonalization for 3D rotation. \cite{gao2020path} proposes a neural representation of position and motion to explain the emergence of grid cell pattern. However, \cite{gao2020path} only considers motion in 2D space and does not take visual input into consideration. Our method can be seen as a generalization of \cite{gao2020path}. Our method models both position and orientation and their corresponding changes in 3D environments, and we associate position representations with visual inputs. The concept of position embedding is also used in other areas such as natural language processing. For example, transformer-based models such as BERT \cite{devlin2018bert} or GPT \cite{radford2018improving} have a high dimensional embedding of the position of word in the sentence. These embeddings \cite{vaswani2017attention, shaw2018self, gehring2017convolutional, wang2021position} can be either learnable or predefined. We introduce learnable representations for camera pose in 3D vision. Our rotation loss enforces translation invariance, which serves as a regularization on the learned representations.   

\subsection{Novel view synthesis}
Learning neural 3D scene representation is a fundamental problem in 3D vision, and a compelling way to evaluate the learned representations is by novel view synthesis. One line of work \cite{sitzmann2019scene, mildenhall2020nerf, tung2019learning} incorporates prior knowledge of rendering such as rotation and projection to enforce the consistency between different views, such as NeRF \cite{mildenhall2020nerf}. Another theme \cite{tatarchenko2016multi, worrall2017interpretable, eslami2018neural} learns neural representations purely from the perception of the agent, without extra 3D prior knowledge. Our model belongs to the latter. Different from previous methods, we also learn neural representations of the camera pose and camera movement, and the representations of 3D scene and camera pose are disentangled in an unsupervised manner. \cite{tatarchenko2016multi, worrall2017interpretable} infer the scene representation from a single image or a pair of images, while \cite{eslami2018neural} assumes that the representation can be obtained from a small batch of images. Compared to these methods, our model is able to utilize posed images of various scenes to update the shared camera pose representations. 


\subsection{Interpretable representation}
In generative modeling, learning interpretable latent representation is a long-standing target. Specifically, the goal is to learn latent vectors such that each dimension or sub-vector is aligned with an independent factor or concept. This can be done either with supervision \cite{kulkarni2015deep, plumerault2020controlling} or without supervision \cite{higgins2016beta, kim2018disentangling, karras2019style}. Besides vector representation, \cite{litany2020representation, worrall2017interpretable,jayaraman2015learning,gao2019learning} learns matrix representation of image transformation that operates on the latent vector representation.

Our model is a combination of both vector and matrix representations. On the one hand, we disentangle the vector representations of each individual scene and camera pose. On the other hand, we model the movement of the camera pose by matrix representation, which is in the form of matrix Lie group and matrix Lie algebra. In terms of parametrization of the matrix representation, \cite{worrall2017interpretable} uses predefined and fixed rotation matrix, \cite{litany2020representation} learns a fixed matrix for each type of variation, and \cite{jayaraman2015learning} parametrizes 2D ego-motion operated on 2D images. Different from these methods, we parametrize the matrix representation of camera movement as a nonlinear function of the movement in 3D that can take continuous values and operate on the vector representations of 3D scenes.   

\subsection{Deep pose regression models}
Deep pose regression models~\cite{sattler2019understanding} can be categorized into absolute camera pose regression (APR)~\cite{kendall2015posenet, kendall2017geometric, brahmbhatt2018geometry} which directly predicts the camera pose given an input image, and relative camera pose estimation (RPR)~\cite{saha2018improved,balntas2018relocnet, laskar2017camera} that predicts the pose of a test image relative to one or more training images. In this work, we adopt the APR setting while the method can also be easily adapted to the RPR setting. Note that our focus is to compare the effectiveness of different camera pose representations, which is orthogonal to the other methods that specifically target at improving the performance of pose regression.


\section{Representational model} \label{sec: model}

Suppose an agent move in a 3D environment with head rotations. There are at most 6 degrees of freedom (DOF), i.e., the position of the agent $(x, y, z)$ and its head orientation $(\alpha, \beta, \gamma)$. We denote them as the pose of the agent $\vp = (x, y, z, \alpha, \beta, \gamma)$. Following the idea of \cite{gao2020path}, we encode each DOF by a $d$-dimensional sub-vector $\vv_l(l), l \in \{x, y, z, \alpha, \beta, \gamma\}$. From the embedding point of view, essentially we embed the 1D domain in $\R^1$ as a 1D manifold in a higher dimensional space $\R^d$. We limit each sub-vector to have unit length, i.e., we further assume the 1D manifold to be a circle. For notation simplicity, we concatenate those sub-vectors to a pose vector $\vv(\vp)$. When the camera makes a movement $\delta \vp = (\delta x, \delta y, \delta z, \delta \alpha, \delta \beta, \delta \gamma)$, the camera pose changes from $\vv(\vp)$ to $\vv(\vp + \delta \vp)$. See Figure \ref{fig:illustrate idea} for an illustration of our proposed framework. 


\subsection{Modeling movement as vector rotation} \label{sec: model1}

We start from considering an infinitesimal camera movement $\delta \vp$. For each DOF $l \in \{x, y, z, \alpha, \beta, \gamma\}$, we propose the following model:
\begin{equation}
   \bm{v}_l (l + \delta l) = \mM_l(\delta l) \vv_l(l) + \bm{o}(\delta l),
    \label{eq1}
\end{equation}
where $\mM_l(\delta l)$ is a $d \times d$ matrix depending on $\delta l$. Given that $\delta l$ is infinitesimal, the model can be further parametrized as
\begin{equation}
   \bm{v}_l (l + \delta l) = (\bm{I} + \bm{B}_l \delta l) \vv_l(l) + \bm{o}(\delta l),
    \label{eq3}
\end{equation} 
where $\bm{I}$ is the identity matrix and $\bm{B}_l$ is a $d \times d$ matrix that needs to be learned. We assume $\bm{B}_l$ to be skew-symmetric i,e. $\bm{B}_l = -\bm{B}_l^T$. This assumption guarantees that $(\bm{I} + \bm{B}_l\delta l)(\bm{I} + \bm{B}_l\delta l)^T = \bm{I} + \bm{o}(\delta l^2)$, i.e., $(\bm{I} + \bm{B}_l\delta l)$ is approximately an orthogonal matrix. From the geometric perspective, it maps the movement along $l$ axis in 1D space to rotation of the vector in the high-dimensional latent space. In practice, we only need to parametrize the upper triangle of $\bm{B}_l$ as trainable parameters and the lower triangle of $\mB_l$ is taken to be the negative of the upper triangle. We further assume $\bm{B}_l$ to be block-diagonal so that the total number of parameters can be greatly reduced. If there are movements on multiple DOFs, we only need to rotate each sub-vector of DOF independently.

As pointed out by \cite{gao2020path}, equations \ref{eq1} and \ref{eq3} can be justified as a minimally simple recurrent model. To model the movement in the latent space, the most general form is $\bm{v}_l (l + \delta l) = F(\bm{v}_l(l), \delta l)$, i.e., the pose vector for the new pose is a function of the one for the old pose and the movement. Given that $\delta l$ is infinitesimal, we can apply the first-order Taylor expansion:
 $   \bm{v}_l(l + \delta l) = \bm{v}_l(l) + f(\bm{v}_l(l)) \delta l + \bm{o}(\delta l),$
where we use $f(\bm{v}_l(l))$ to denote the first derivative of $F(\bm{v}_l(l), \delta l)$, i.e., $f(\bm{v}_l(l)) = \frac{\partial}{\partial \delta l} F(\bm{v}_l(l), \delta l) |_{\delta l = 0}$. When the movement $\delta l = 0$, we should have that $F(\bm{v}_l(l), 0) = \bm{v}_l(l)$. Then a minimally simple model is to assume $f(\bm{v}_l(l))$ as a linear transformation , i.e. $f(\bm{v}_l(l)) = \bm{B}_l\bm{v}_l(l)$, and $\bm{v}_l(l + \delta l) = \bm{v}_l(l) + \bm{B}_l\bm{v}_l(l) \delta l + o(\delta l)$. As we will discuss in \ref{small change approximate}, for finite movement $\Delta \vp$, we recurrently apply the model of infinitesimal $\delta \vp$, so that the matrix representation becomes a matrix Lie group.

\begin{figure}[t!]
    \centering
    \includegraphics[width=0.45\textwidth]{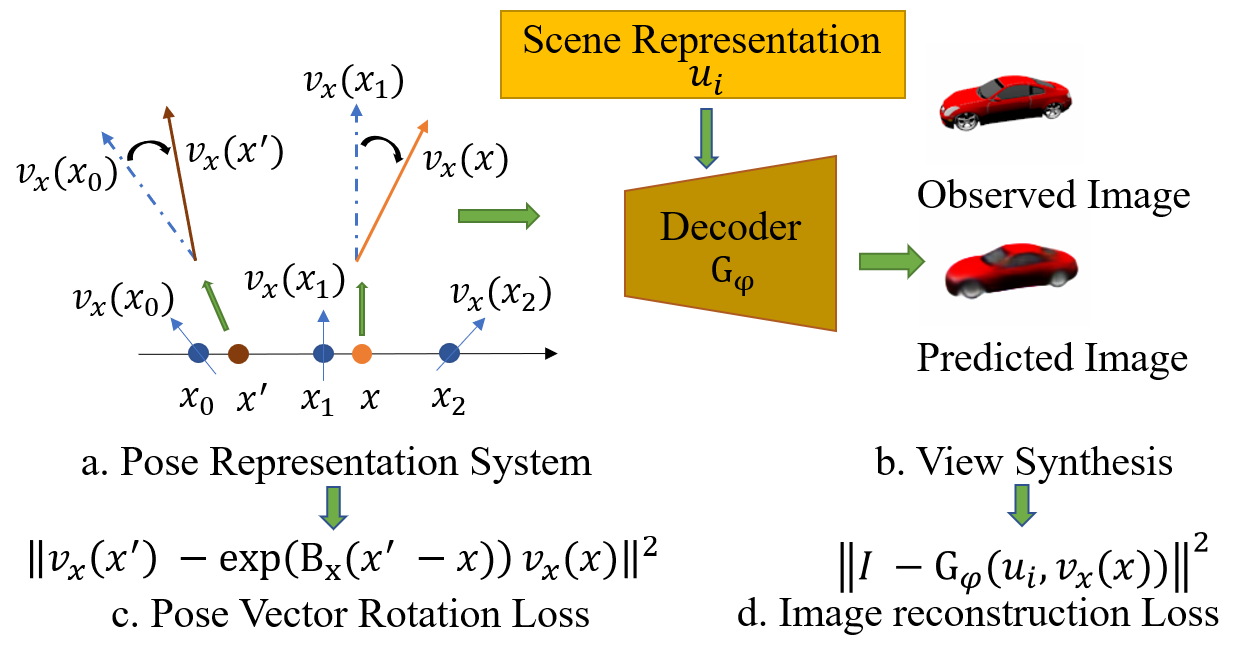}
    \caption{\small Illustration of our framework. (a) Pose vector for a given position $x$ is obtained by rotating its nearby grid vector. (b) Pose vector is fed-in a decoder together with a scene representation vector to predict image under certain view. (c) The rotation consistency of our pose representation system is enforced through pose rotation loss. (d) The decoding ability of our pose representation system is enforced through image reconstruction loss.}
    \label{fig:illustration frame-work}
\end{figure}

\subsection{Polar  system for position change in 2D}
If the movement of the agent is constrained in a 2D environment, we follow \cite{gao2020path} to use a polar coordinate system to model the change of position, which corresponds to the egocentric perspective and could be potentially more biological plausible. Specifically, let $\vx = (x, y)$ be the position of the agent in the 2D space, instead of using individual vector $\bm{v}_x$ and $\bm{v}_y$ to represent the position, we represent position in a single vector $\bm{v}_{\vx}(\vx)$ and the movement is captured by direction $\theta$ and distance $\delta r$. We have $\delta \vx = (\delta x, \delta y) = (\delta r \cos \theta, \delta r \sin \theta)$. The representational model under this polar coordinate system is:
\begin{equation}
    \bm{v}_\vx(\vx + \delta \vx) =  (\bm{I} + \bm{B}(\theta)\delta r)\bm{v}_\vx(\vx) + \bm{o}(\delta r).
    \label{eqn4}
\end{equation}
The $\bm{B}(\theta)$ is a function of $\theta$ and is skew-symmetric. $\bm{B}(\theta)$ models the change of position along the direction $\theta$. If the agent changes the direction of movement from $\theta$ to $\theta + \delta \theta$, then we assume  
\begin{equation}
    \bm{B}(\theta + \delta \theta) = (\bm{I} + \bm{C}\delta\theta)\bm{B}(\theta) + \bm{o}(\delta\theta),
    \label{eqn5}
\end{equation}
where $\bm{C}$ is another skew-symmetric matrix to learn. The geometric interpretation is that if the agent changes direction, $\bm{B}(\theta)$ is rotated by another matrix $\bm{C}$.  

For camera movement in 3D environment, such coupled representation in polar coordinate will end up with too many matrix representations to learn. Therefore, we restrict ourselves in using it only in 2D space, and use the vector-matrix representations that are disentangled for each DOF as proposed in \ref{sec: model1} for general 3D movements.

\subsection{Matrix Lie group for finite movement}
\label{small change approximate}
So far we have discussed the formulation for infinitesimal movements above. In this subsection we generalize to finite movements. Suppose the agent has a finite movement $\Delta l$ along the axis $l \in \{x, y, z, \alpha, \beta, \gamma\}$. We can divide this movement into $N$ steps, so that as $N \rightarrow \infty$, $ \frac{\Delta l}{N} \rightarrow 0$, and  
\begin{align}
    \bm{v}_l(l + \Delta l) &= 
    (\bm{I} + \bm{B}_l(\frac{\Delta l}{N}) + \bm{o}(\frac{1}{N}))^N \bm{v}_l(l) \nonumber \\
    &\rightarrow \exp(\bm{B}_l\Delta l) \bm{v_l}(l).
    \label{eqn6}
\end{align}
This underlies the relationship between matrix Lie algebra and matrix Lie group. Specifically, the set of $\mM_l(\Delta l) = \exp(\mB_l\Delta l)$ for $\Delta l \in \R$ forms a matrix Lie group. The tangent space of $\mM(\Delta l)$ at identity is the corresponding matrix Lie algebra. $\mB_l$ is the basis of this tangent space, and is also called as the generator matrix. 

For a finite but small $\Delta l$, $\exp(\mB_l\Delta l)$ can be approximated by a second-order Taylor expansion
\begin{equation}
\exp(\bm{B}_l\Delta l) = 
\bm{I} + \bm{B}_l\Delta l + \frac{1}{2} \bm{B}_l^2 \Delta l^2  + \bm{o}(\Delta l^2).
\label{eqn7}
\end{equation}
For a large finite change in each axis, we can divide it into a series of small finite changes, expand each change using second-order Taylor expansion and multiply them together.

\subsection{Theoretical understanding of our model}
A deep theoretical result from mathematics, namely the Peter-Weyl theorem~\cite{taylor2002lectures},  inspires our work. It says that for a compact Lie group, if we can find an irreducible unitary representation, i.e., each element $\vx$ of the group is represented by a unitary (or orthogonal) matrix $\bm{M}(\vx)$, then the matrix elements ($M_{ij}(\vx)$) form a set of orthogonal basis functions for the general functions of $\vx$. This is a deep generalization of Fourier analysis. In our case, the learned vector representation $\vv(\vx) = \bm{M}(\vx) \vv(0)$ are linear compositions of the above basis functions, and the elements ($v_i(\vx)$) serve as a more compact set of basis functions for representing general functions of $\vx$. Our method can be used to represent the pose of the camera and objects in general. The continuous changes of the pose in the physical space generally form a Lie group. Our learned vector and matrix system forms a representation of the pose and its change in the neural space.

\subsection{Implementation of pose representation} \label{sect:implementation}
Suppose we want to learn the representation of axis l, whose value ranges in $[a, b]$. For orientation, the angles is of range [\ang{0}, \ang{360}), while for position, we can predefine the largest range the agent can move within. We divide this range into multiple grids and we learn an individual vector at each grid point. Given an arbitrary position $l \in [a, b]$, we first find its nearest grid point and the corresponding vector representation, and then we rotate this vector to the target position by the matrix representation depending on the distance between this nearest grid position and the target position. See Figure \ref{fig:illustration frame-work}. Since we can set the length of grid to be relatively small,  the distance between the grid and target positions is also small, so that we can use second-order Taylor expansion in Equation \ref{eqn7} to approximate the matrix representation. 

\subsection{Decoding to posed 2D images}
To associate the camera pose representation with visual input, more specifically the posed 2D images, we propose a decoder or emission model. For each 3D scene, suppose we are given multiple posed 2D images $\rmI$ and the corresponding camera poses $\vp$. Then we assume a shared vector representation $\vu$ of the 3D scene, and obtain the vector representation of the camera pose $\vv(\vp)$ as described in \ref{sect:implementation}. We learn a decoder $G_\phi$ that maps $\vu$ and $\vv(\vp)$ to the image space to reconstruct $\rmI$
\begin{equation}
    \hat{\rmI} = G_\phi(\vu, \vv(\vp)),
\end{equation}
where $\phi$ denotes parameters in the decoder network. 

\section{Learning and inference}
\subsection{Learning through view synthesis}
\label{sec: gen}
For a general 3D environment, the unknown parameters of the proposed model include (1) $\vv(\vp)$ for any $\vp$ on grid positions, (2) $\mB_l$ for any $l \in \{x, y, z, \alpha, \beta, \gamma\}$, and (3) parameters $\phi$ in $G_\phi$. To learn these parameters, we define a loss function $L = \lambda_1 L_{\rm rec} + \lambda_2 \sum_{l \in \{x, y, z, \alpha, \beta, \gamma\}} L_{{\rm rot}, l}$, where
\begin{align}
    L_{\rm rec} & = \E_{\rmI} \norm{\rmI - G_\phi(\vu, \vv(\vp))}^2, \nonumber \\
    L_{{\rm rot}, l} & = \E_{l, \Delta l} \norm{\bm{v}_l(l + \Delta l) - \exp(\bm{B}_l(\Delta l))\bm{v}_l(l)}^2.
      \label{eqn9}
\end{align}
$L_{\rm rec}$ is the reconstruction loss for view synthesis, which enforces the decoding of the pose and scene representations to reconstruct the observation. The expectation is estimated by Monte Carlo samples. $L_{\rm rot}$ stands for rotation loss, which serves to constrain $\vv_l$ so that the learned pose representations of different poses can be transformed to each other based on our representational model (Equation \ref{eqn6}). The expectation term in $L_{\rm rot}$ can be approximated by randomly sampled pairs of poses $\vp$ and $\vp + \Delta \vp$ that are relatively close to each other, which means that we have infinite amount of data for this loss term. 

If the movement of camera pose is in a 2D space and we employ the polar coordinate system, then part (2) of the unknown parameters becomes $\mB_l$ for any $l \in \{\alpha, \beta, \gamma\}$, $\mB(\theta)$ and $C$. The loss functioin is defined as $L = \lambda_1 L_{\rm rec} + \lambda_2 \sum_{l \in \{\alpha, \beta, \gamma\}} L_{{\rm rot}, l} + \lambda_3 L_{{\rm rot}, \vx} + \lambda_4 L_{{\rm rot}, \theta}$, where $L_{\rm rec}$ and $L_{{\rm rot}, l}$ follow equation \ref{eqn9} and
\begin{align}
L_{{\rm rot}, \vx} &= \E_{\vx, \Delta \vx} \norm{\bm{v}_\vx(\vx + \Delta \vx) - \exp(\mB(\theta)\Delta r)\bm{v}_\vx(\vx)}^2, \nonumber \\
    L_{{\rm rot}, \theta} &= \E_{\vx} \norm{\bm{B}(\theta + \Delta \theta) - \exp(\bm{C}\Delta \theta))\bm{B}(\theta)}^2.
    \label{eqn9.2} 
\end{align}

For training, we minimize $L$ by iteratively updating the decoder $G_\phi$ (as well as our scene representation $\bm{u}$) and our pose representation system $v_l$, $M_l$ for $ l \in \{x, y, z, \alpha, \beta, \gamma\}$. In practice, the decoder is parameterized by a multi-layer deconvolutional neural network. Besides the latent vector on top of the decoder, we also learn a scene-dependent vector at each following layers using AdaIN \cite{huang2017arbitrary}. We normalize the scene vector at the top layer of the decoder to have unit norm so that it has the same magnitude as the pose representation. We find this helps optimization. More details can be found in Supplementary.

\subsection{Inference by pose regression}
\label{sec: infer}
With the learned pose representation, we can then use it as the target output for camera pose regression. Specifically, for each DOF, we train a separate inference network $E_{\xi l}$ that maps the observed posed 2D image $\rmI$ to the pose representation $\vv_l(l)$ . The loss function is defined as the $L_2$ distance between the inferred and learned pose presentations 
\begin{equation}
    L_l = \E_{\rmI} \norm{\vv_l(l) - E_{\xi l}(\rmI)}^2.
\end{equation}
In practice, $E_{\xi l}$ is parameterized by a convolutional neural network where $\xi$ denotes the parameters and we introduce some scene-dependent parameters using AdaIN. For different DOFs, the inference networks share common lower layers but with different top fully-connected layers. 

For testing, given an unseen posed image $\rmI$, we can get the inferred pose representation $\bm{\hat{v}_l}$ from our inference model, and decode the predicted pose by:
\begin{equation}
    \hat{l} = \underset{l}{\arg\min}\norm{\bm{v_l}(l) - \bm{\hat{v}_l}} ^2, \; l \in (x, y, z, \alpha, \beta, \gamma)
    \label{eqn12}
\end{equation}

\section{Experiments}

In this section, we demonstrate the efficacy of our learned pose representations in both view synthesis and pose regression tasks. For view synthesis, we mainly compare with the Generative Query Network(GQN) \cite{eslami2018neural}.  For pose regression, we compare our learned neural representations of camera pose with other commonly used pose representations, including the Euler angle, the sinusoidal representation used in GQN, and the quaternions (as well as log quaternions) representations used in the PoseNet \cite{kendall2015posenet, kendall2017geometric} and MapNet \cite{brahmbhatt2018geometry}, by evaluating the pose estimation accuracy. More details of implementation can be found in Supplementary. Our code and pretrained models can be found at \url{https://github.com/AlvinZhuyx/camera_pose_representation}.

\subsection{Datasets}
\textbf{GQN rooms}. GQN~\cite{eslami2018neural} introduces a synthetic dataset with 2 million synthetic scenes, where each scene contains various objects, textures, and walls. The agent can navigate in a 2D space and rotate the head horizontally in the scenes. Each scene contains 10 rendered $64 \times 64$ RGB images. We use the version of the dataset where the camera moves freely and the objects do not rotate around their axes. We sample $200,000$ scenes from the dataset. For each scene, we sample 9 images for training and use the left one image for testing. Since this dataset contains a large number of simple scenes with a small number of images for each scene, instead of learning an individual scene representation vector for each scene, we learn an encoder to encode the scene representation online similar to \cite{eslami2018neural}. Since the agent has 2 DOFs for position and 1 DOF for orientation, our pose vector contains one position sub-vector in the polar coordinate system and one orientation sub-vector. Each sub-vector has 96 dimensions. We assume that $B$ is block-diagonal with six blocks, and each block is 16 $\times$ 16. 

\textbf{ShapeNet v2}. We use the images generated by \cite{sitzmann2019scene} from the \textit{car} category of ShapeNet v2 dataset~\cite{chang2015shapenet}. This dataset contains 2,151 object models. For each scene, the instance locates at the center of a sphere. The virtual agent can move on the surface of this sphere, with its camera pointing to the center. Therefore, the agent has 2 DOFs, and we use 2 orientation angles to denote its position on the sphere. Each instance contains 500 different views of $128 \times 128$ rendered RGB images, where we randomly sample 100 images for training and leave the others for testing. The pose representation contains two sub-vectors of two orientation angles. Each sub-vector has a dimension of 96, and $B$ has six $16 \times 16$ blocks. We learn an individual scene representation vector $\vu$ for each instance. 

\textbf{Gibson Environment}. The Gibson Environment \cite{xia2018gibson} provides tools for rendering images corresponding to different views in a room, which we use to generate a synthetic dataset. We refer to this dataset as Gibson rooms. Specifically, we select 20 areas of size 2m $\times$ 2m from different rooms. For each area, we randomly render about 28k $128 \times 128$ RGB images of different views. We fix the camera height and constrain the camera to rotate only horizontally. Compared to GQN rooms and ShapeNet car, this synthetic dataset contains more realistic and complicated indoor scenes, which could be more challenging. Moreover, it includes fewer scenes while for each scene, images from abundant views are provided. Therefore, incorporating view-based information becomes very important. The agent has 2 DOFs for position and 1 DOF for orientation, which corresponds to a position sub-vector in the polar coordinate system and one orientation sub-vector. The dimensions of the sub-vectors and $B$ are the same as the ones for GQN rooms dataset.   

\textbf{7 Scenes Dataset}.
Microsoft 7 Scenes~\cite{shotton2013scene} is a widely used dataset for camera pose estimation. It contains RGB-D images for seven different indoor scenes. Each scene has several trajectories for training and testing. In our experiment, we follow the training and testing split in~\cite{shotton2013scene}, and we only use RGB images without depth information. We translate and align the position coordinates of scenes and ensure that all the trajectories locate in a 4m $\times$ 1.5m $\times$ 3m cuboid. The agent has 6 DOFs, so the pose representation vector contains 6 sub-vectors. We assume that each sub-vector has a dimension of 32, and each $B$ has four 8 $\times$ 8 blocks. We mainly use this dataset for camera pose regression. We resize the images to 128 $\times$ 128 when training the decoder and pose representation system. We use shared pose representations for all the seven scenes and distinct scene representation for each of them. When performing pose regression, following \cite{kendall2017geometric, brahmbhatt2018geometry}, we train an individual inference model for each scene and resize the input images so that the shortest side is of length 256.  

\subsection{Novel view synthesis}

\begin{figure}[t!]
\centering
\begin{subfigure}{.32\linewidth}
    \centering
    \includegraphics[width=0.99\textwidth]{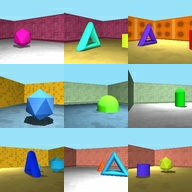}
\end{subfigure}
    \hfill
\begin{subfigure}{.32\linewidth}
    \centering
    \includegraphics[width=0.99\textwidth]{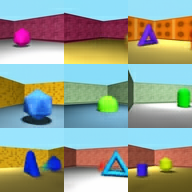}
\end{subfigure}
   \hfill
\begin{subfigure}{.32\linewidth}
    \centering
    \includegraphics[width=0.99\textwidth]{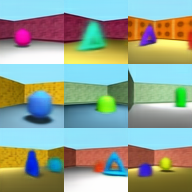}
\end{subfigure}

\begin{subfigure}{.32\linewidth}
    \centering
    \includegraphics[width=0.99\textwidth]{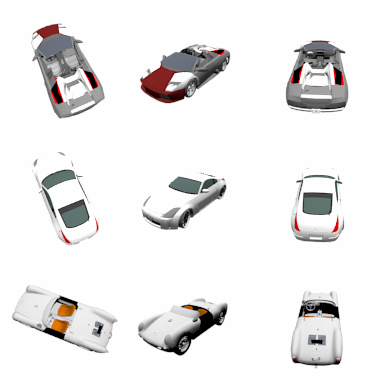}
\end{subfigure}
    \hfill
\begin{subfigure}{.32\linewidth}
    \centering
    \includegraphics[width=0.99\textwidth]{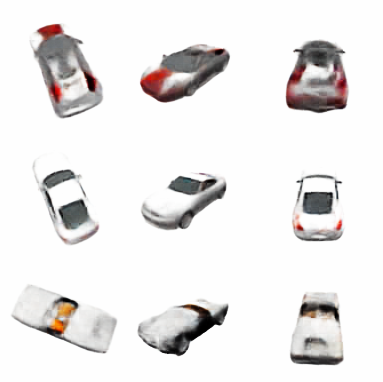}
\end{subfigure}
   \hfill
\begin{subfigure}{.32\linewidth}
    \centering
    \includegraphics[width=0.99\textwidth]{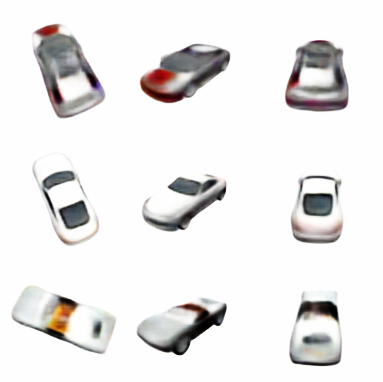}
\end{subfigure}

\begin{subfigure}{.32\linewidth}
    \centering
    \includegraphics[width=0.99\textwidth]{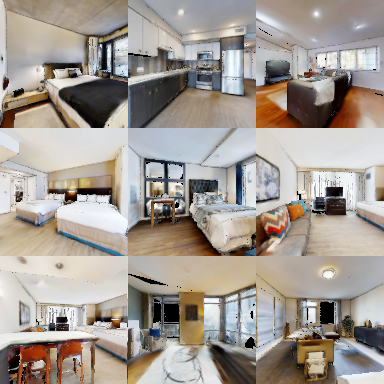}
    \caption{\small Ground Truth}\label{fig:image31}
\end{subfigure}
    \hfill
\begin{subfigure}{.32\linewidth}
    \centering
    \includegraphics[width=0.99\textwidth]{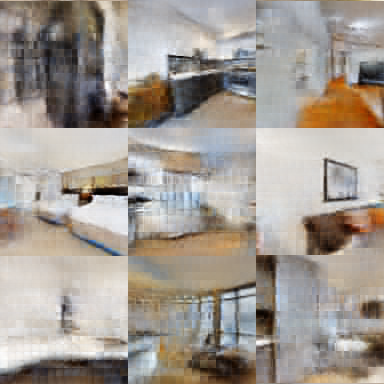}
    \caption{\small GQN}\label{fig:image32}
\end{subfigure}
   \hfill
\begin{subfigure}{.32\linewidth}
    \centering
    \includegraphics[width=0.99\textwidth]{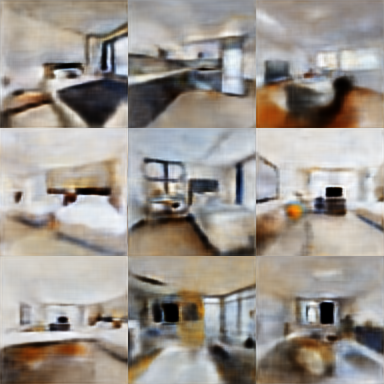}
    \caption{\small Ours}\label{fig:image33}
\end{subfigure}

\caption{\small Qualitative results for novel view synthesis. {\em Top}: GQN rooms. {\em Middle}: ShapeNet car. {\em Bottom}: Gibson rooms.}
\label{fig:novel view}
\end{figure}

\begin{figure*}[t]
\centering
\begin{subfigure}{.28\linewidth}
    \centering
    \includegraphics[width=0.99\textwidth]{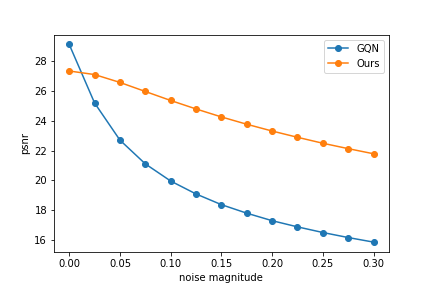}
    \caption{\small GQN rooms}\label{fig:image211}
\end{subfigure}
    \hfill
\begin{subfigure}{.28\linewidth}
    \centering
    \includegraphics[width=0.99\textwidth]{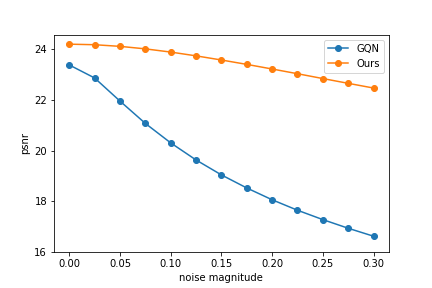}
    \caption{\small ShapeNet car}\label{fig:image212}
\end{subfigure}
   \hfill
\begin{subfigure}{.28\linewidth}
    \centering
    \includegraphics[width=0.99\textwidth]{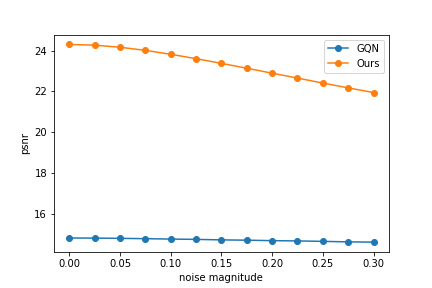}
    \caption{\small Gibson rooms}\label{fig:image213}
\end{subfigure}
\caption{\small Quantitative results for novel view synthesis given different noise magnitudes. In each figure, we plot the PSNR over different magnitudes of noise introduced to the position vector. For a given noise magnitude $\alpha$, if the $i$-th element in the position vector has a standard deviation $\beta_i$, then we add a Gaussian Noise $N(0, (\alpha\beta_i)^2)$ to the corresponding element. Noise magnitude 0.0 corresponds to the novel view synthesis test result. We compare with GQN on three datasets.} 
\label{fig:psnr}
\end{figure*}

The first question is whether our learned pose representation is meaningful. We answer this by testing our learned representations on novel view synthesis task. The experimental results demonstrate that our learned representations can generate a novel view of a scene of high quality. Figure \ref{fig:novel view} shows the qualitative results, and Figure \ref{fig:psnr} shows the quantitative results in terms of Peak Signal-to-Noise Ratio (PSNR). We compare the results with GQN. For GQN, we use the implementation by \cite{GQN:2018} and the same training and testing splits as ours. We use 8 generation layers and set the shared core option to be False. We add extra convolution and de-convolution layers when dealing with images of size $128 \times 128$. The total number of parameters for this GQN implementation is 114M. In contrast, our model only has less than 9M parameters. 

From Figure \ref{fig:novel view} and Figure \ref{fig:psnr} (noise magnitude of 0.0 corresponds to novel view synthesis test result), we see that for GQN rooms dataset, our model gets a bit worse but comparable results with the GQN model. For ShapeNet car dataset, which contains complex instances, our model generates more consistent and clearer results compared with GQN. For Gibson rooms dataset, which is more complicated, GQN fails to capture the relationship. The reconstruction only captures some specific views and does not generalize to other views. On the other hand, our learned model is able to generate a query view corresponding to our pose representation. This is probably because that the 3D scene representations in our method are learned by all the 2D posed images of the scene. 
 
 \begin{table*}[t]
    \centering
    \begin{tabular}{llllllll}
        \toprule
        \multirow{2}{*}{Representations} & ShapeNet car & \multicolumn{3}{c}{GQN rooms} & \multicolumn{3}{c}{Gibson rooms}\\
        \cmidrule{2-2} \cmidrule{3-5} \cmidrule{6-8} 
        {} & orientation & \; $x$ & \; $y$   & orientation & \;\; $x$ &\; \; $y$ & orientation \\
        \midrule
        $(x, y, \alpha)$ &  \ang{7.75} & 0.069 & 0.071 &  \ang{12.00} & 0.043m & 0.041m & \ang{7.03}\\
        ($x$, $y$, axis-angle) & \ang{11.29} & \;\; - & \; \;- & \;\; - & \; \;- & \;\; - & \; \;- \\
        ($x$, $y$, $\sin(\alpha)$, $\cos(\alpha)$)) &  \ang{7.29} & 0.108 & 0.104 &  \ang{16.46} & 0.033m & 0.034m &  \ang{1.19}\\
        ($x$, $y$, $q$) & \ang{4.28} & \textbf{0.050} & \textbf{0.048} & \ang{5.34}  & 0.043m & 0.042m &  \ang{1.21}\\
        ($x$, $y$, $\log q$) & \ang{5.35} & 0.051 & 0.051 &  \ang{7.44}  & 0.028m & 0.027m &  \ang{1.17}\\
        ours &  \textbf{\ang{2.85}} & 0.053 & 0.053 &  \textbf{\ang{4.07}} & \textbf{0.021m} & \textbf{0.020m} &  \textbf{\ang{0.87}}\\
        \bottomrule
    \end{tabular}
    \caption{\small Camera pose estimation errors on different datasets. We compare with several camera pose representations. ($x$, $y$, $\alpha$) denotes the representation that uses $x, y, z$ coordinate to represent position and Euler angle to represent orientation. ($x$, $y$, axis-angle) denotes using axis-angle representation for rotation. Note that for GQN rooms and Gibson rooms datasets, the agent only has one DOF of rotation. Therefore, the axis-angle representation degrades to one Euler angle representation, and its results should be the same as the Euler angle. ($x$, $y$, $\sin(\alpha), \cos(\alpha)$) denotes using sinusoidal functions to represent orientation. ($x$, $y$, $q$) denotes the unit quaternions representation used in \cite{kendall2017geometric} while ($x$, $y$, $\log q$) stands for the logarithm quaternions representation proposed in  \cite{brahmbhatt2018geometry}. Our method uses learned pose vectors for both camera position and orientation. We report the average prediction error for each dataset. For ShapeNet car dataset, the camera is located on a sphere, so we only need to predict the orientation angle. For GQN rooms and Gibson rooms datasets, we predict both the camera position and orientation. For GQN rooms, the range of each scene is from -1.0 to 1.0. For Gibson rooms, we render each scene to an area of 2m $\times$ 2m.}
    \label{table:infer}
\end{table*}
 
 \subsection{Robustness to pose noise}
Next, we try to answer why we need that representation and what is the advantage of such neural representation over directly using 6 DOFs coordinate representation in terms of novel view synthesis. One critical supporting evidence is that our learned neural pose representation is more robust to noise. Specifically, Figure \ref{fig:psnr} shows the changes of PSNR for our model versus the GQN model when some Gaussian noise with various magnitudes is added to the pose representations. We observe that the performance of the GQN model degrades quickly as the magnitude of added noise increases. This is not surprising since GQN directly uses coordinate representation for position and orientation and thus is vulnerable to noise interference. On the other hand, our learned representation embeds the camera pose to high dimensional space and is further regulated by the rotation loss, and thus is more robust to noise.

\begin{table*}[t]
    \centering
    \begin{tabular}{llllll}
    \toprule
        Scene & PoseNet17\cite{kendall2017geometric} & PoseNet + $\log q$\cite{brahmbhatt2018geometry} & PoseNet + $\log q$ (*) & ours \\
        \midrule
        Chess & 0.13m, \ang{4.48} & \textbf{0.11m}, \textbf{\ang{4.29}} &  0.17m \ang{4.96} & 0.12m \ang{4.83} \\
        Fire & \textbf{0.27m}, \ang{11.30} & \textbf{0.27m}, \ang{12.13} & 0.36m \ang{11.22} & \textbf{0.27m}
        \textbf{\ang{8.91}} \\
        Heads & 0.17m, \ang{13.00} & 0.19m, \textbf{\ang{12.15}} & 0.20m \ang{13.35} & \textbf{0.16m} \ang{12.84} \\
        Office & \textbf{0.19m}, \textbf{\ang{5.55}} & \textbf{0.19m}, \ang{6.35} & 0.23m \ang{7.05} & \textbf{0.19m} \ang{6.64} \\
        Pumpkin & 0.26m, \textbf{\ang{4.75}} & 0.22m, \ang{5.05} & 0.26m \ang{5.87} & \textbf{0.22m} \ang{5.45}\\
        Red Kitchen & \textbf{0.23m}, \ang{5.35} & 0.25m, \textbf{\ang{5.27}} & 0.29m \ang{6.10} & 0.24m \ang{6.10} \\
        Stairs & 0.35m, \ang{12.40} & 0.30m,\ang{11.29} & 0.36m \textbf{\ang{10.18}} & \textbf{0.29m}  \ang{10.70} \\
        Average & 0.23m, \ang{8.12} & 0.22m \ang{8.07} & 0.27m \ang{8.39} & ${\textbf{0.21m}}$ ${\textbf{\ang{7.92}}}$ \\
        \bottomrule
    \end{tabular}
    \caption{\small Camera pose estimation errors on 7scenes dataset. We compare our results with PoseNet using quaternions (PoseNet17) and log quaternions (PoseNet + $\log q$). The column PoseNet + $\log q$(*) are the results we get by running the code provided by \cite{brahmbhatt2018geometry}. In the last column, we show the results using our learned pose representation. Following the convention, we report the median prediction errors here.}
    \label{tab:7Scenes}
\end{table*}

 \subsection{Inference results}
 We further demonstrate that our learned representation is efficient to serve as the target output of pose regression. In the camera pose regression task, the camera position is usually represented using 3D coordinate $(x, y, z)$ and the camera orientation can be represented by various methods. The most straightforward one is to use the Euler angle to represent the orientation. Another representation is axis-angle representation. In \cite{eslami2018neural}, the authors use $(\sin(\alpha), \cos(\alpha))$ to represent each orientation angle. Besides, unit quaternions and logarithm of the unit quaternions are another two popular representations used in pose regression \cite{kendall2015posenet, kendall2017geometric, brahmbhatt2018geometry}. Comparing with those methods, we used learned neural representations for both camera position and orientation. We conduct the pose regression experiments on all four datasets we mentioned above. For our representation, Euler Angle representation and $(\sin(\alpha), \cos(\alpha))$ representation we use mean square error loss for regression. On 7 Scenes dataset, we also use $L_1$ norm loss for our representation. For quaternions and log quaternions representations, as suggested by \cite{brahmbhatt2018geometry}, we use $L_1$ norm loss. For axis-angle representation, we find that for ShapeNet car dataset, using $L_1$ norm loss leads to better results. For Gibson rooms and GQN rooms datasets, since the agent can only rotate its head horizontally, the axis-angle representation degrades to a single Euler angle. For the two quaternions-related baselines, we employ the automatic weight tuning method proposed in \cite{kendall2017geometric} to make a fair comparison. Note that the main focus of this work is to compare different pose representations, and thus we do not include other improvement techniques (\eg, including unlabeled data or relative pose loss between image pairs), as we consider them as orthogonal directions to improving the pose representations. More details can be found in Supplementary.  
 
 We first show the comparison results on GQN rooms, ShapeNet car, and Gibson rooms datasets in Table \ref{table:infer}. For a fair comparison, we keep the same network structure for all the representations on each dataset and only change the final output layer. Since the dimension of our learned representation is higher than all the baseline representations, for a fair comparison, we add another fully-connected layer to these baseline inference networks so that the inference models have roughly the same number of parameters across different pose representations. According to Table \ref{table:infer}, our representation consistently outperforms all the other representations, especially for orientation regression. For most configurations, our representation yields the best results in both orientation and position prediction. On GQN dataset, the quaternions and log quaternions representation achieve slightly better results in position prediction. However, their orientation prediction results are much worse than ours. A possible explanation is that we embed both the camera position and orientation as neural representations, and thus they are more consistent with each other. Besides, representing the rotation angles on a hyper-sphere in a high dimensional space may also make it easier for the model to regress. 
 
 We further compare our learned pose representations with the popular quaternions and log quaternions representations on 7 Scenes dataset using PoseNet. Following \cite{brahmbhatt2018geometry}, we use a pre-trained ResNet34 as our feature extractor and 6 parallel fully-connected (FC) layers to predict the 6 pose sub-vectors. We employ color jittering as data augmentation and remove the dropout in the FC layers. The results are shown in Table \ref{tab:7Scenes}. We compare our results with \cite{kendall2017geometric, brahmbhatt2018geometry}. We also run the code provided by \cite{brahmbhatt2018geometry} to re-train their model and report the results. The difference between the reported values and the reproduced results is probably due to the randomness and different versions of software \footnote{The code of \cite{brahmbhatt2018geometry} is originally implemented in python 2.7 and PyTorch 0.4.0 while we make minor adaptation to enable it to run in python 3.6 and PyTorch 1.2.0}. Following the convention on this dataset, we report the median errors of location and orientation predictions. The result shows that, on average, our model outperforms all the baselines. 

\section{Conclusion and Future Work}
We propose a framework for learning neural vector representations for both camera poses and 3D scenes, coupled with neural matrix representation for camera movements. The model is learned through novel view synthesis and can be used for camera pose regression. Our learned representation proves to be more robust against pose noise in the novel view synthesis task and works well as the estimation target for camera pose regression. We hope that our work can motivate further interest and study on learning neural representations for camera poses and joint representations for camera poses and 3D scenes. An interesting future direction is how to combine our method with the recent work of NeRF \cite{mildenhall2020nerf}, which uses sinusoidal functions of very high frequencies. Our model can be adapted to this new generative model structure and may be able to learn more flexible camera pose representation. 

{ \section*{Acknowledgment}

The work is supported by NSF DMS-2015577;  DARPA XAI  N66001-17-2-4029; ARO  W911NF1810296; ONR MURI  N00014-16-1-2007.}

{\small
\bibliographystyle{ieee_fullname}
\bibliography{egbib}
}

\appendix
\title{Appendix}
   
\maketitle
\pagestyle{empty}

\section{Training details}
In this section, we describe the details about the structure of our neural networks and the hyperparameters we use in the experiments. The main differences among the network structures we use on different datasets depend on: (i) the size of the image we are dealing with: the larger image needs more blocks; (ii) the complexity of the scenes. For 7Scenes and Gibson rooms dataset, the scenes are highly complex. Therefore we apply instance normalization to multiple layers, which is dependent on scenes, besides the vector representation of the scene at the top layer. For the GQN rooms dataset, which includes a huge amount of scenes, we employ an encoder to calculate the scene representations online. We use Adam\cite{kingma2014adam} as optimizer for all the experiments with $\beta_1 = 0.9$ and $\beta_2 = 0.999$. The learning rate for each setting is introduced in each later section.

\subsection{GQN rooms dataset}
\begin{figure*}[ht!]
\centering
\begin{subfigure}{.45\linewidth}
    \centering
    \includegraphics[width=0.99\textwidth]{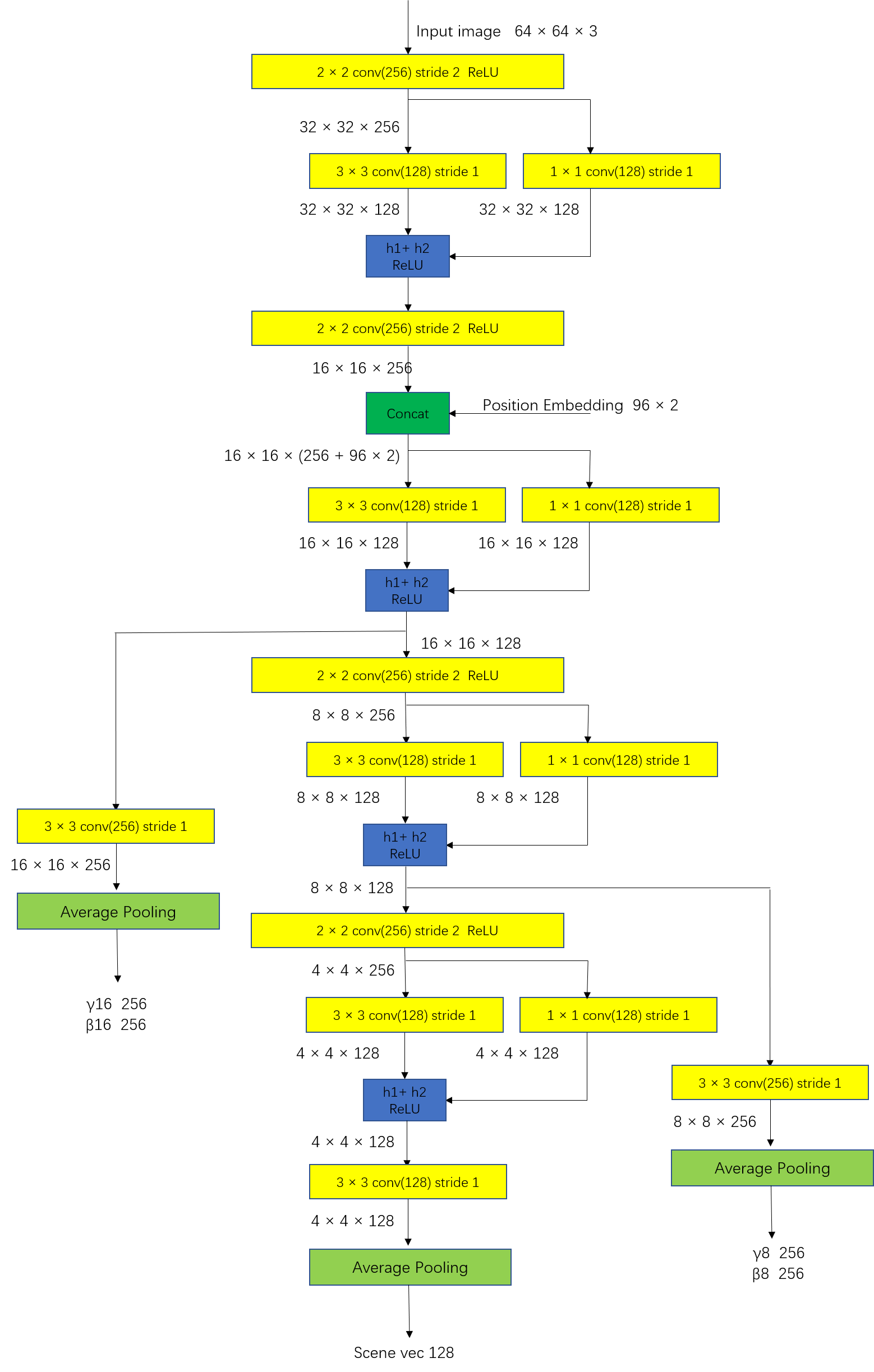}
    \caption{\small Scene Encoder}
    \label{fig:GQN encoder}
\end{subfigure}
    \hfill
\begin{subfigure}{.29\linewidth}
    \centering
    \includegraphics[width=0.99\textwidth]{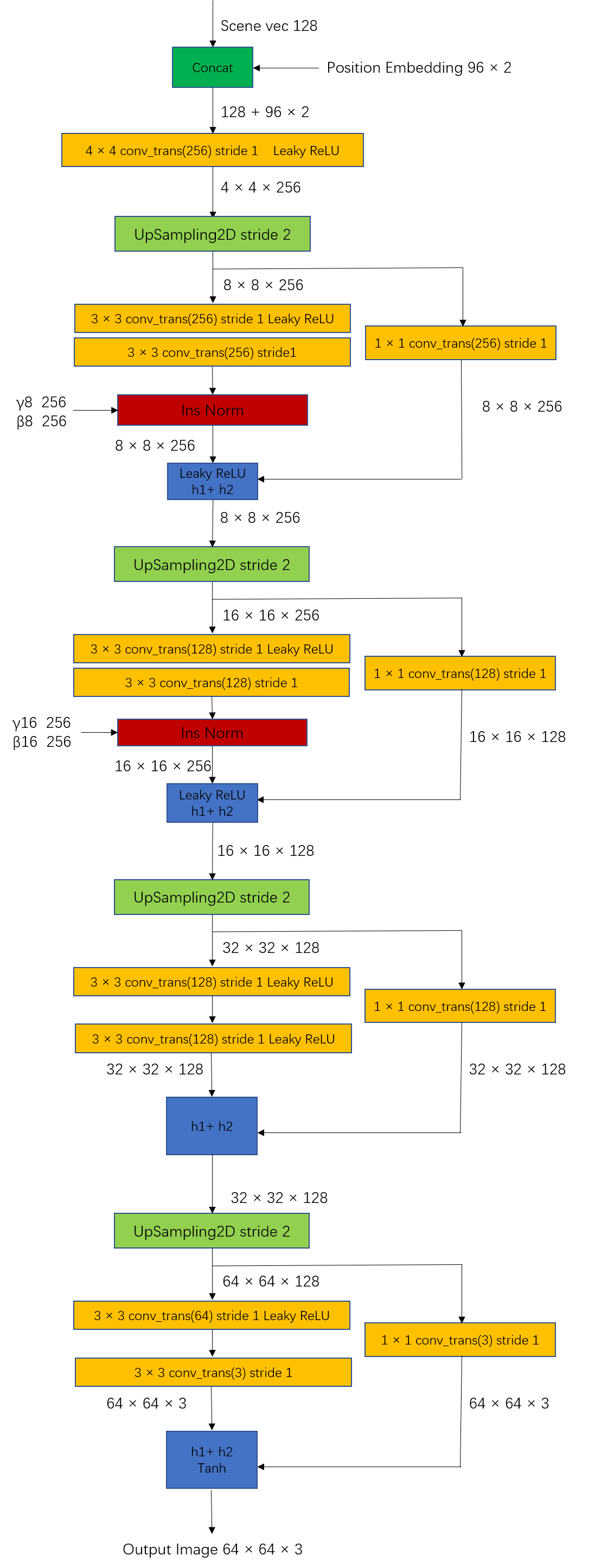}
    \caption{\small Decoder}
    \label{fig:GQN decoder}
\end{subfigure}
   \hfill
\begin{subfigure}{.25\linewidth}
    \centering
    \includegraphics[width=0.99\textwidth]{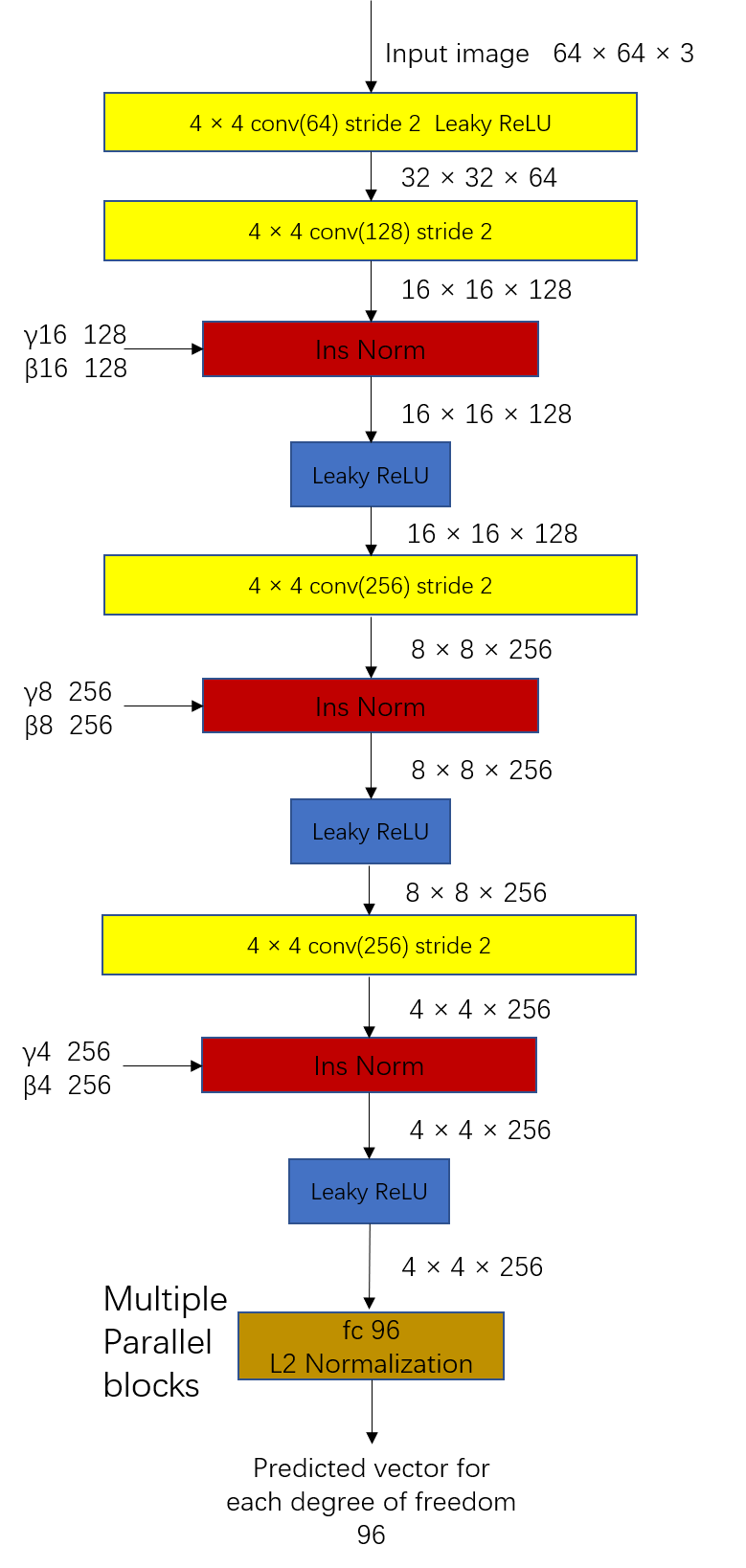}
    \caption{\small Inference model}
    \label{fig:GQN inference}
\end{subfigure}
\caption{\small Network structures for GQN rooms dataset. $\gamma$ and $\beta$ denote the parameters of instance normalization.}
\label{fig:GQN room}
\end{figure*}

\textbf{Generative experiment}. Since this dataset contains a huge amount of scenes, and each scene only has few images, we encode the scene representations online instead of learning an individual vector representation for each scene. The encoder structure is shown in Figure \ref{fig:GQN encoder}. Specifically, the encoder encodes the scenes as a scene vector that is fed to the top layer of the decoder, and it also encodes the parameters of instance normalization \cite{huang2017arbitrary} that is applied to the multiple layers of the generator. Following \cite{eslami2018neural}, to summarize information across multiple images of the same scene, we sum up the encoded vectors and parameters of these images. The decoder structure is shown in Figure \ref{fig:GQN decoder}. We discretize the square space into 20 $\times$ 20 grids and learn a position vector at each grid. Similarly, we discretize the orientation into 36 grids (\ang{10} per grid) and learn an orientation vector at each grid. The training takes about four days on a single Titan RTX GPU.

We train the model for one million iterations. At each iteration, we randomly sample 30 scenes, each containing ten images. We use the first six images of each scene to encode the scene representation and concatenate it with the other three images' pose representations. We use the concatenated representations to reconstruct the three images. We leave the last image for testing. For the rotation loss, we randomly sample 4000 pairs of poses for each iteration. The learning rate of the pose representations and matrix representations of camera movements is 0.01, and the learning rate for the encoder, decoder, and scene representations is 0.0001. Here, we update all the learnable parameters together. We set $\lambda_1$ as 0.05, $\lambda_2$, $\lambda_3$ as 100 and $\lambda_4$ as 0.8.

For the baseline GQN network, we also train the model for one million steps. At each step, we feed in a batch of 64 scenes. The other parameters follow the original implementation. 

\textbf{Inference experiment}. We show the inference model structure in Figure \ref{fig:GQN inference}. Like the generative experiment, we use an encoder to encode the scene and the parameters of instance normalization online. The encoder structure is the same as the encoder used in the generation task, except that we do not encode a vector representation at the top layer but encode another set of instance norm parameters $(\gamma4, \beta4)$. We set the learning rate as 0.0001 for all the parameters. We train the inference model for 100,000 steps. At each iteration, we feed in 30 scenes. For this dataset, we use the homoscedastic uncertainty method proposed in \cite{kendall2017geometric} to automatically tune the weight between pose prediction loss of position and orientation. We set the initial guess for logarithmic weight of position loss as $S_{\rm pos} = -\log{20}$ and the initial guess for logarithmic weight of orientation loss as $S_{\rm ori} = -\log{5}$ (so that $\exp(-S_{\rm pos}) = 20$ and $\exp(-S_{\rm ori}) = 5$). We use the same inference model structure for baseline models, except that we add another fully-connected (FC) layer with size 196 to these models to make sure that they have approximately the same amount of parameters as the model trained on our representations. We also train these models for 100,000 iterations with the same batch size. We tune the learning rate for each baseline model to make a fair comparison and use the same automatically weight tuning method for the two quaternions-related baselines. The initial guess for the logarithm weight of position loss is set to 0.0, and the one of orientation loss is set to -3.0 as suggested by \cite{kendall2017geometric}.  For our model and each of the baseline models, the training takes about 5 hours on a single Titan RTX GPU.

\subsection{ShapeNet car}
\begin{figure*}[ht!]
\centering
\begin{subfigure}{.55\linewidth}
    \centering
    \includegraphics[width=0.99\textwidth]{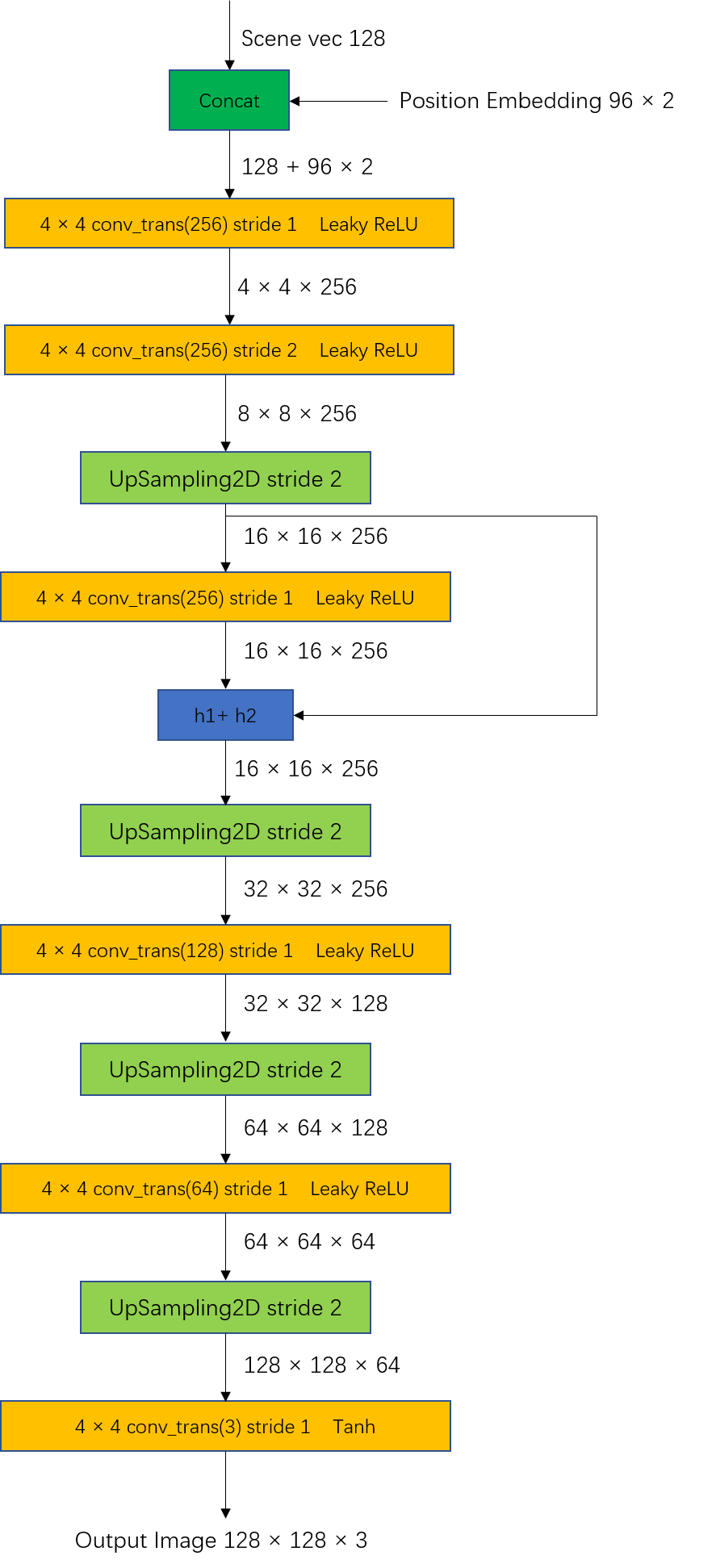}
    \caption{\small Decoder}
    \label{fig:Car decoder}
\end{subfigure}
    \hfill
\begin{subfigure}{.43\linewidth}
    \centering
    \includegraphics[width=0.93\textwidth]{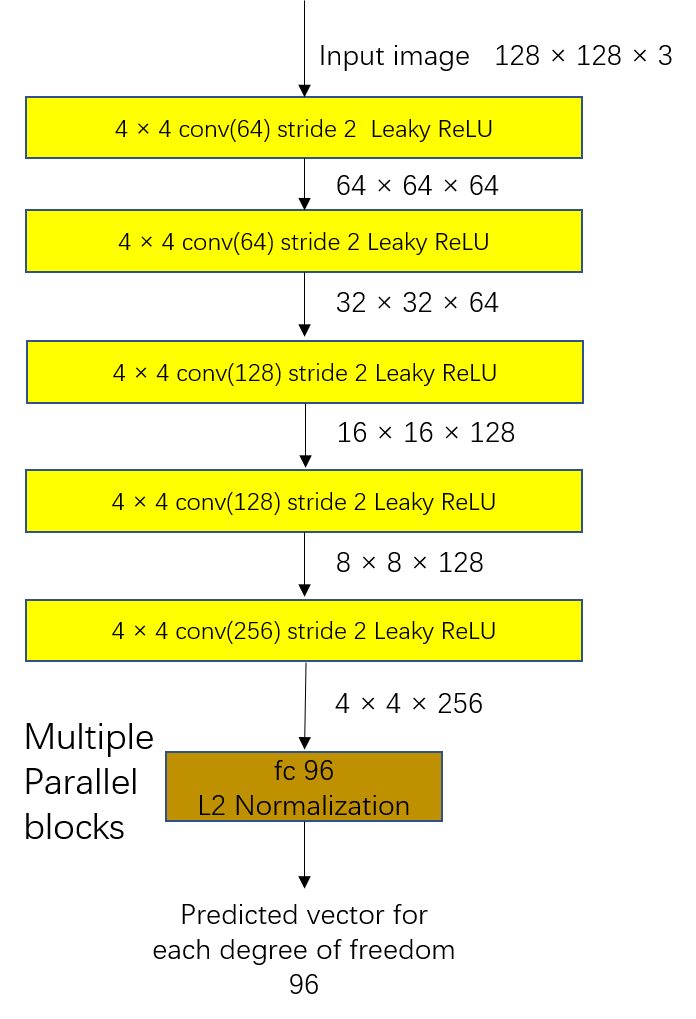}
    \caption{\small Inference}
    \label{fig:Car inference}
\end{subfigure}

\caption{\small Network structures for ShapeNet car dataset.}
\label{fig:Car}
\end{figure*}

\textbf{Generative experiment}. This dataset contains 2151 different cars. The heads of the cars are aligned to the same orientation, and the background is blank. Given the simplicity of this dataset, we do not use instance normalization. The vector representation of scenes is of 128 dimensions, and we learn a separate vector representation for each scene instead of obtaining by an encoder. The structure of the generator model is shown in Figure \ref{fig:Car decoder}. For our pose representation system, we discretize the orientation for \ang{0} to \ang{360} into 36 grids and learn individual orientation vectors at each grid.

For each scene, we randomly sample 50 pairs of images for each scene as the training set and leave the others as the test set. The camera poses of the two images in each pair is close to each other, so that the change from one to another can be approximated by Taylor expansion of the matrix Lie groups as discussed in section 3.3, which means that we can apply the camera poses of the two images to the rotation loss. We train our model for 160,000 iterations, i.e., 1500 epochs. We randomly sample 20 scenes at each iteration, and for each instance, we sample 10 images (5 pairs). For the rotation loss, we randomly sample additional 200 pairs of camera poses to compute the loss. The learning rate is set to 0.0001. We set $\lambda_1$ as 0.05 and $\lambda_2$ as 50. We iteratively update the decoder for one time and pose representation system for three times at each iteration. The training takes about four days on a single Titan RTX GPU.

For the baseline GQN model, we trained the model for 500,000 steps. At each step, we randomly sample a batch of 36 scenes. We randomly sample 15 images for each scene to infer the scene representation and another image as the reconstruction target. We use the same train-test split as our model for each scene here.

\textbf{Inference experiment} Since the head direction for each car is aligned to the same direction, the pose regression task should follow the same rule across different scenes. Thus, we do not include scene-related parameters in our inference model. The structure of our inference model is shown in Figure \ref{fig:Car inference}. For each scene, we randomly sample 250 images as the training set and the rest 250 images as the test set. We train our model and all the baseline models for 500 epochs. At each iteration, we use 10 scenes, and we randomly sample 20 images from each scene. The learning rate is set to 0.001. We simply set the weights of prediction losses of the two rotation vectors as 1.0 without further automatic tuning. For each baseline representation, we use the same inference model structure and add another fully-connected (FC) layer with size 256. We tune the learning rate carefully to make a fair comparison, and we use the automatic weight tuning method for the two quaternions-related baseline methods. The initial guess for the logarithmic weight of orientation loss is set to -3.0 as suggested by \cite{kendall2017geometric}. The training for our model and each of the baseline models takes about 8 hours on a single Titan RTX GPU. 

\subsection{Gibson rooms dataset}
\begin{figure*}[ht!]
\centering
\begin{subfigure}{.55\linewidth}
    \centering
    \includegraphics[width=0.75\textwidth]{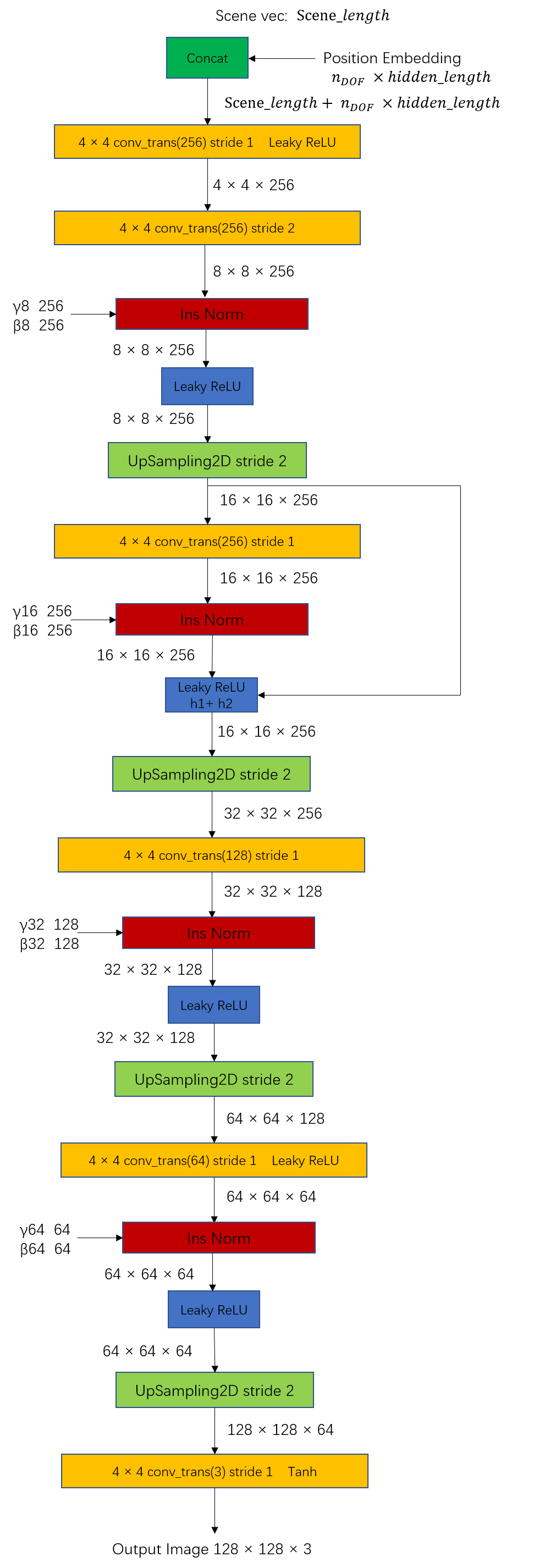}
    \caption{\small Decoder}
    \label{fig:real room decoder}
\end{subfigure}
    \hfill
\begin{subfigure}{.43\linewidth}
    \centering
    \includegraphics[width=0.9\textwidth]{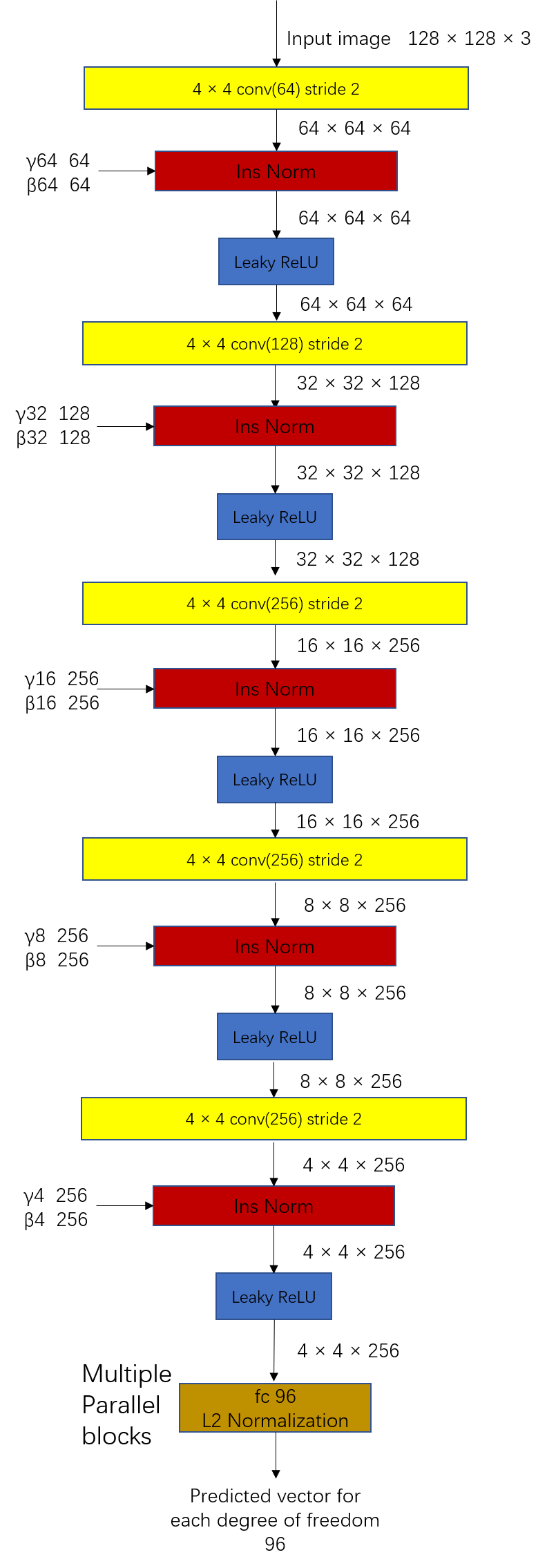}
    \caption{\small Inference}
    \label{fig:real room inference}
\end{subfigure}
\caption{\small Network structures for Gibson rooms dataset. $\gamma$ and $\beta$ denote the parameters of instance normalization. The 7Scenes dataset shares the same decoder structure with Gibson room dataset while its inference model is the same as \cite{brahmbhatt2018geometry}.}
\label{fig:Car}
\end{figure*}

\textbf{Generative experiment}.
This dataset contains complex scenes. We apply instance normalization at multiple layers, which is dependent on the scene. The structure is shown in Figure \ref{fig:real room decoder}. The scene vector representation is of 768 dimensions, and the dimensions of instance normalizations are summarized in Figure \ref{fig:real room decoder}. We discretize the 2m $\times$ 2m square space into 40 $\times$ 40 grids. We discretize the two orientation angles into grids so that each grid is \ang{10}. 

For each scene, we randomly sample half of the data as the training set and the rest as the test set. We train our model for 500k steps. At each iteration, we randomly choose four scenes. For each scene, we randomly sample 50 images. For the rotation loss, we randomly sample another 3000 pairs of poses. We use a learning rate of 0.0001 for training the generator and a learning rate of 0.01 for the pose representation. We iteratively update the generator parameters for one time and update the pose representation two times at each iteration. We set $\lambda_1$ as 0.01, $\lambda_2$, $\lambda_3$ as 100 and $\lambda_4$ as 0.8. The training takes about five days on a single Titan RTX GPU.

For the baseline GQN model, we train the model for 500k steps. At each iteration, we randomly sample and predict 36 images. To predict each image, we randomly pick 15 images from the same scene to infer the scene representations.

\textbf{Inference experiment}.
The inference structure is shown in Figure \ref{fig:real room inference}. We trained the inference model for 25000 steps for both our representation and the baseline representations. At each step, we randomly sample 4 scenes with 50 images from each scene. The learning rate for the model with our representation is 0.001. For this dataset, we find that simply set the weight of position prediction loss as 20 and set the weight of orientation prediction loss as 10 is good enough. So we do not employ the automatic weight tuning mechanism here. We tuned the learning rate for each baseline model, and we applied the homoscedastic uncertainty method to tune the weight for the quaternions-related representations automatically. The initial guess for the logarithm weight of position loss is set to 0.0, and the one of orientation loss is set to -3.0. For each baseline representation, we use the same inference model structure and add another fully-connected (FC) layer with size 192.The initial guess follows \cite{kendall2017geometric}. For our model and each of the baseline models, the training takes about 5.5 hours on a single Titan RTX GPU.

\subsection{7Scenes}
\textbf{Generative experiment}.
For this dataset, we use the same generator structure as for the Gibson Room dataset (see Figure \ref{fig:real room decoder}). Since this dataset contains less data than the Gibson Room dataset, we set the dimension of the scene vector representation to 96. We discretize the whole region (4m $\times $ 1.5m $\times$ 3m) into grids so that each grid is of 0.1m $\times$ 0.1m $\times$ 0.1m. The orientation is discretized into grids so that each grid is of \ang{10}.

We update the model for 100,000 steps. At each step, we randomly sample 16 pairs of images from each scene, and we randomly sample 3000 extra pairs of poses to estimate the rotation loss. We use the learning rate 0.0001 for the generator and 0.001 for the pose representation system. We iteratively update the generator for one time and pose representation system for two times at each iteration. We set $\lambda_1$ as 0.009 and $\lambda_2$ as 50. The training takes about one day on a single Titan RTX GPU.

\textbf{Inference experiment}.
For the inference model, we use the same structure proposed in \cite{brahmbhatt2018geometry}, \ie, we use a pre-trained ResNet34 as the basic feature extractor. We learn a separate module containing several FC layers on the top of the extracted features to predict each pose vectors.  Following \cite{brahmbhatt2018geometry}, we train an individual inference model for each scene. We use learning rate 0.00005 and train the model of each scene for 60 epochs. To isolate the effect of different representations, we use PoseNet as the model for all the representations, without other techniques such as adding pair losses or unlabeled data. We consider these techniques to be orthogonal to the improvement in pose representation. We employ the automatic weight tuning method as \cite{brahmbhatt2018geometry} to tune the weight between the three position vectors and three orientation vectors. We set the initial guess for the logarithm weight of three position vectors' losses the initial guess for logarithm weight of three orientation vectors' losses as -3.0. We employ 0.7 color jitter as data augmentation and remove the dropout in the final FC layer. Our model takes about 3.7 hours for training all the 7 scenes on a single Titan RTX GPU. For the baseline model, we use the released code of \cite{brahmbhatt2018geometry} and we use the default setting with python 3.6 and torch 1.2.0, which trains the models on each scene for 300 epochs with a learning rate of 0.0001. It takes about 13 hours to train the baseline models on the entire 7 scenes on a single Titan RTX GPU.

\section{Additional training results}
\subsection{Generative results}
\begin{figure*}[ht!]
\centering
\begin{subfigure}{.32\linewidth}
    \centering
    \includegraphics[width=0.99\textwidth]{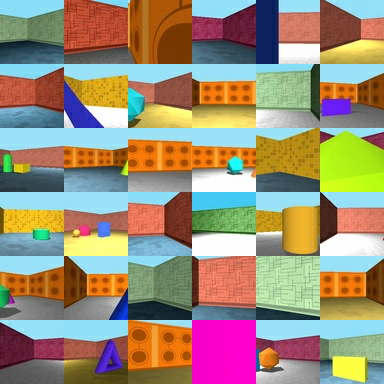}
\end{subfigure}
    \hfill
\begin{subfigure}{.32\linewidth}
    \centering
    \includegraphics[width=0.99\textwidth]{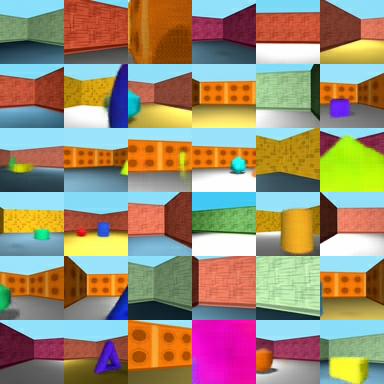}
\end{subfigure}
   \hfill
\begin{subfigure}{.32\linewidth}
    \centering
    \includegraphics[width=0.99\textwidth]{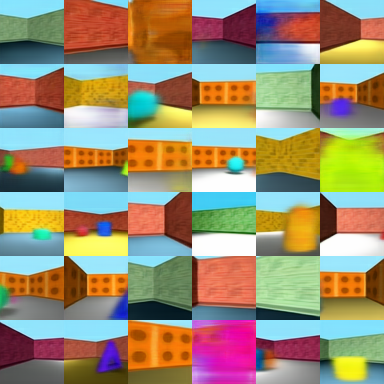}
\end{subfigure}

\begin{subfigure}{.32\linewidth}
    \centering
    \includegraphics[width=0.99\textwidth]{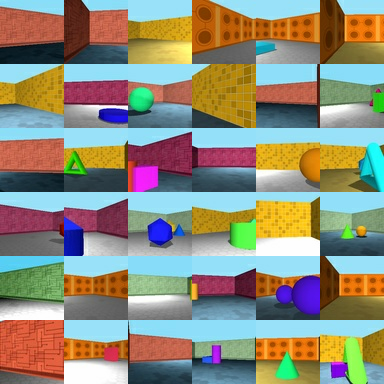}
\end{subfigure}
    \hfill
\begin{subfigure}{.32\linewidth}
    \centering
    \includegraphics[width=0.99\textwidth]{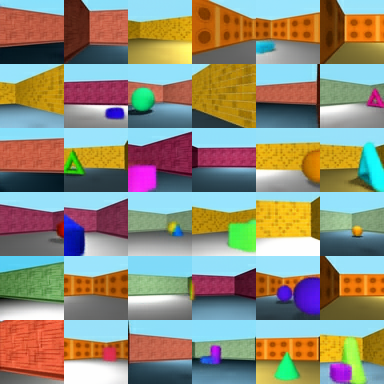}
\end{subfigure}
   \hfill
\begin{subfigure}{.32\linewidth}
    \centering
    \includegraphics[width=0.99\textwidth]{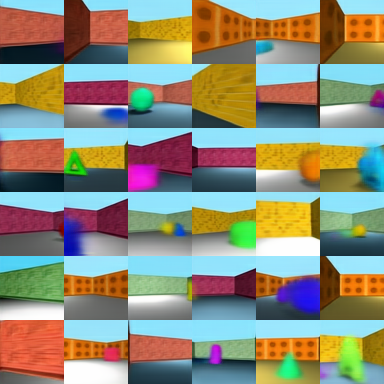}
\end{subfigure}

\begin{subfigure}{.32\linewidth}
    \centering
    \includegraphics[width=0.99\textwidth]{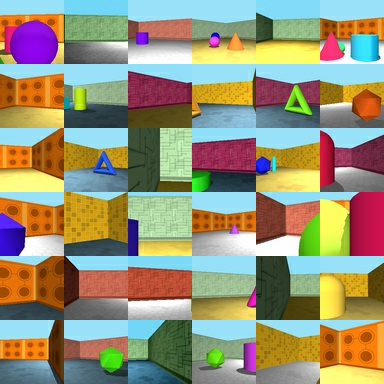}
    \caption{\small Ground Truth}\label{fig:image31}
\end{subfigure}
    \hfill
\begin{subfigure}{.32\linewidth}
    \centering
    \includegraphics[width=0.99\textwidth]{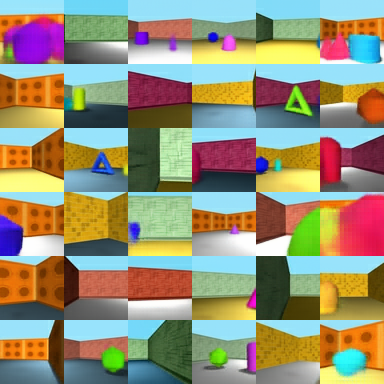}
    \caption{\small GQN}\label{fig:image32}
\end{subfigure}
   \hfill
\begin{subfigure}{.32\linewidth}
    \centering
    \includegraphics[width=0.99\textwidth]{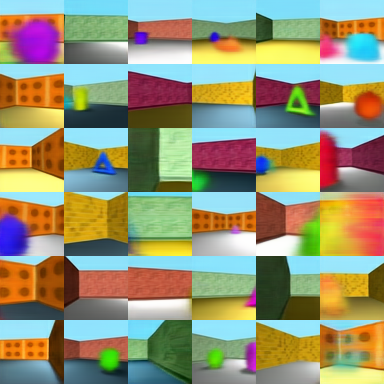}
    \caption{\small Ours}\label{fig:image33}
\end{subfigure}
\caption{\small Additional novel view synthesis results on GQN rooms dataset.}
\label{fig:room}
\end{figure*}

\begin{figure*}[ht!]
\centering
\begin{subfigure}{.32\linewidth}
    \centering
    \includegraphics[width=0.99\textwidth]{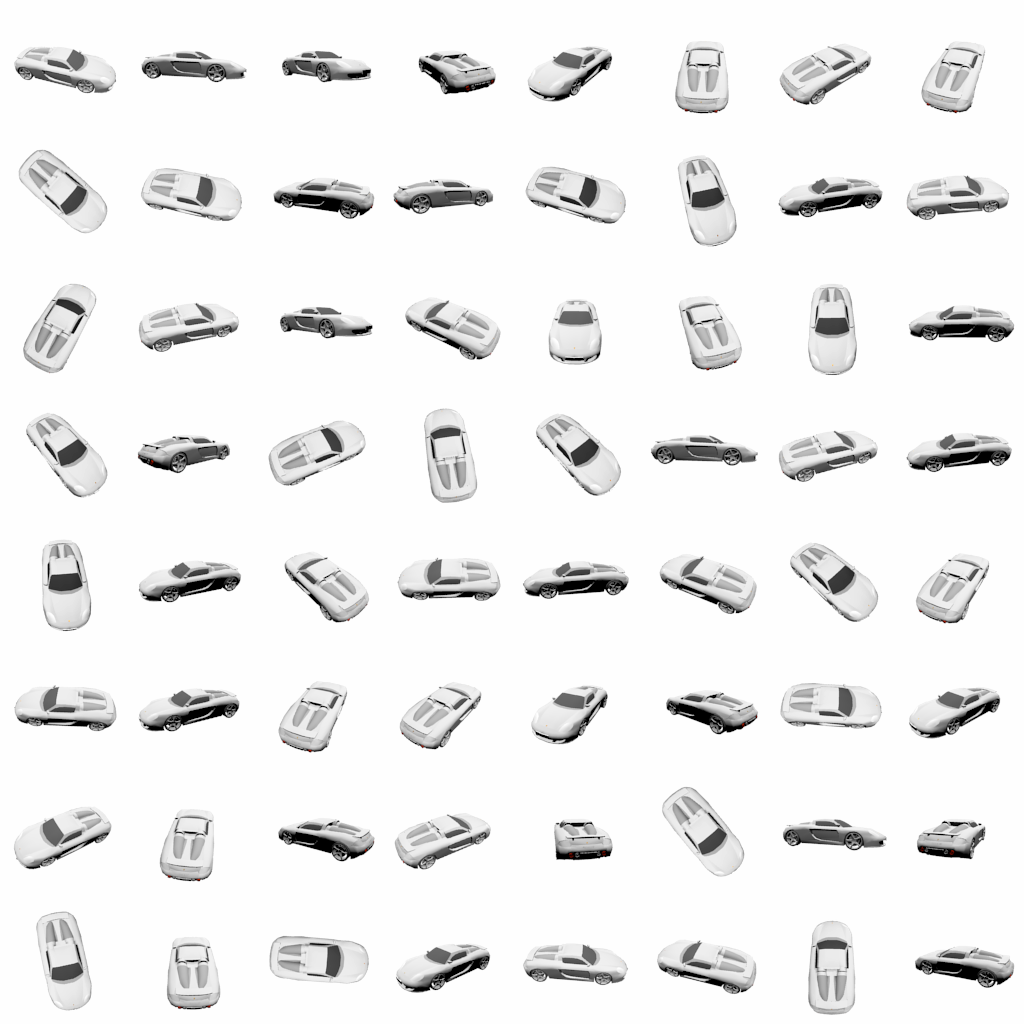}
\end{subfigure}
    \hfill
\begin{subfigure}{.32\linewidth}
    \centering
    \includegraphics[width=0.99\textwidth]{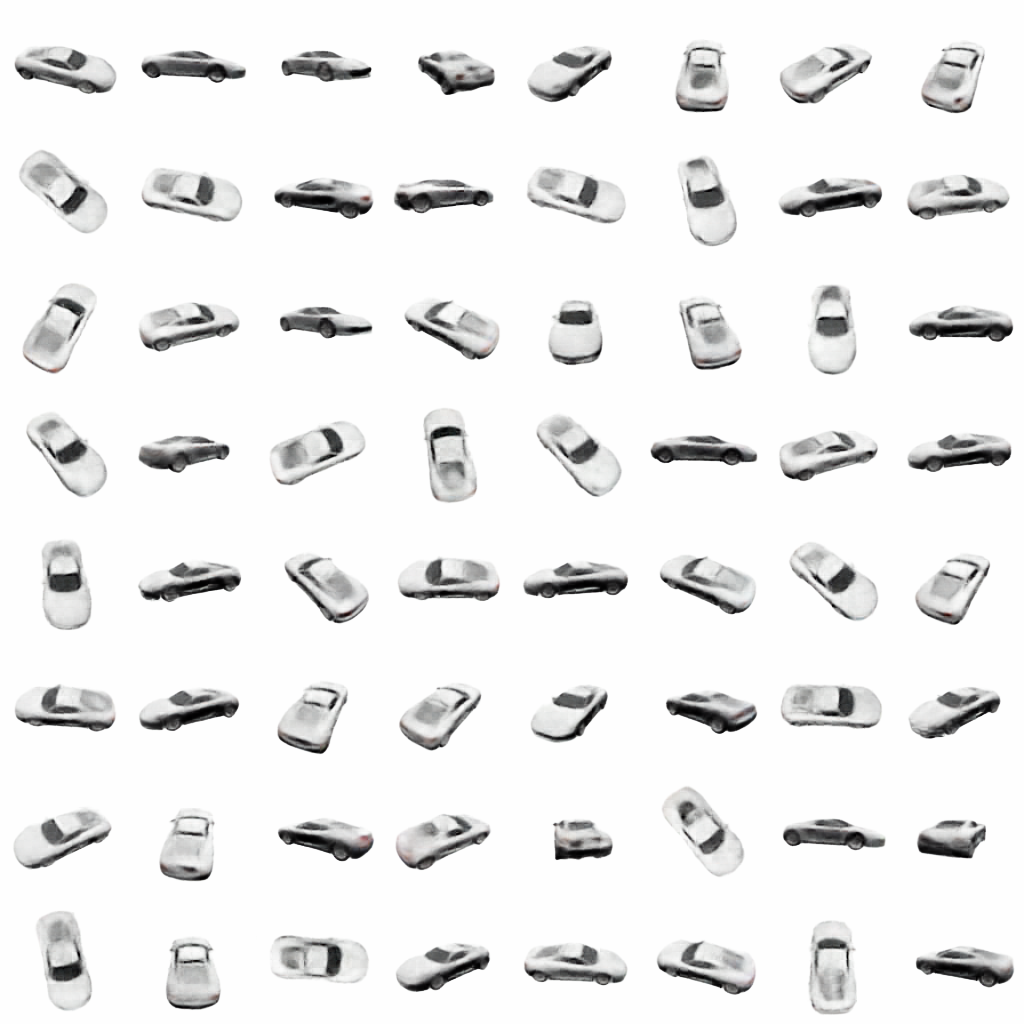}
\end{subfigure}
   \hfill
\begin{subfigure}{.32\linewidth}
    \centering
    \includegraphics[width=0.99\textwidth]{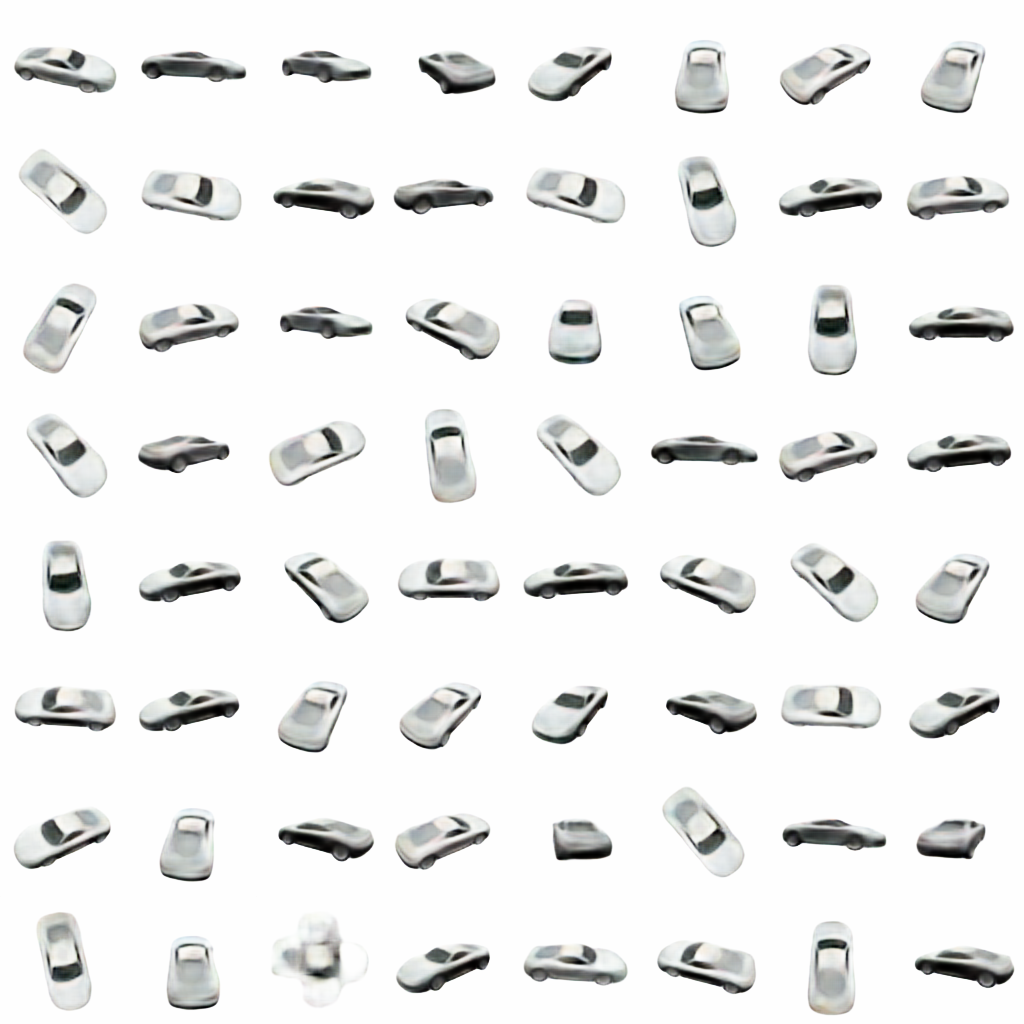}
\end{subfigure}

\begin{subfigure}{.32\linewidth}
    \centering
    \includegraphics[width=0.99\textwidth]{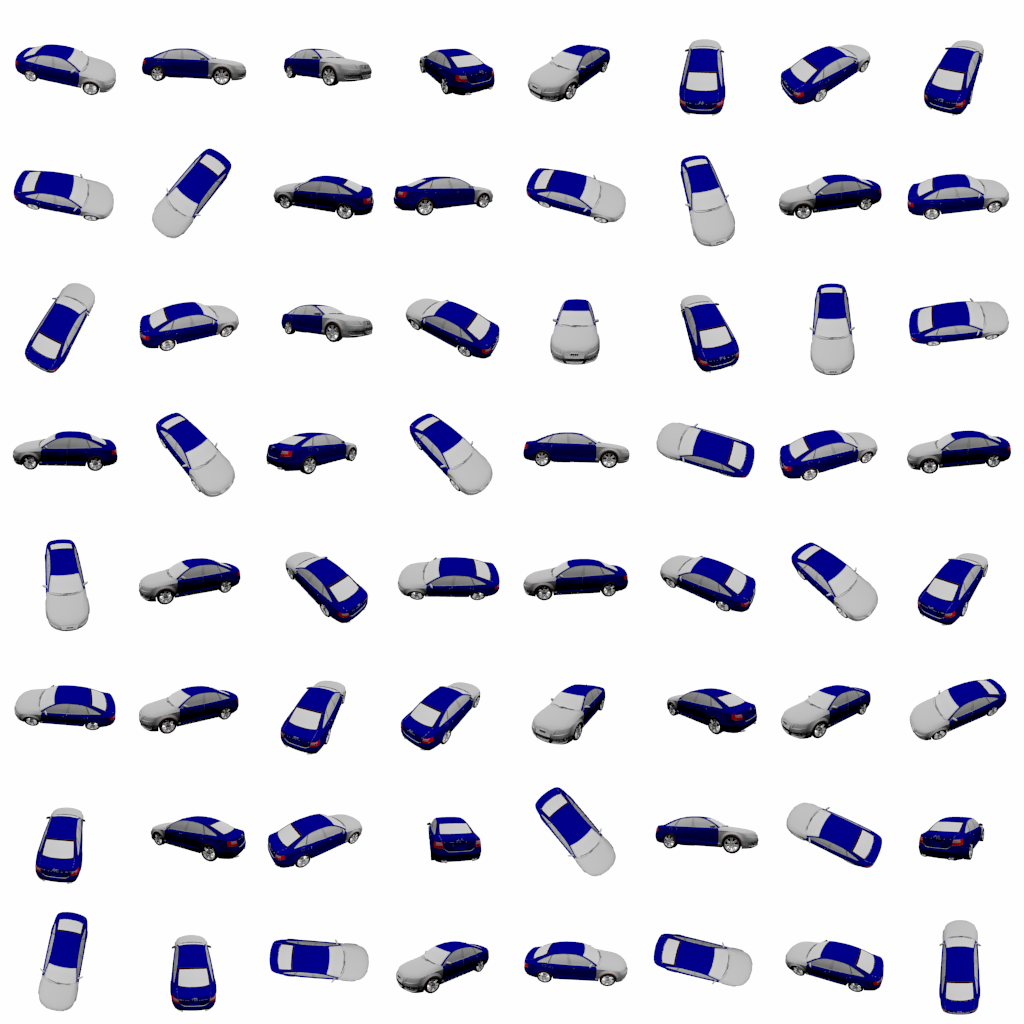}
\end{subfigure}
    \hfill
\begin{subfigure}{.32\linewidth}
    \centering
    \includegraphics[width=0.99\textwidth]{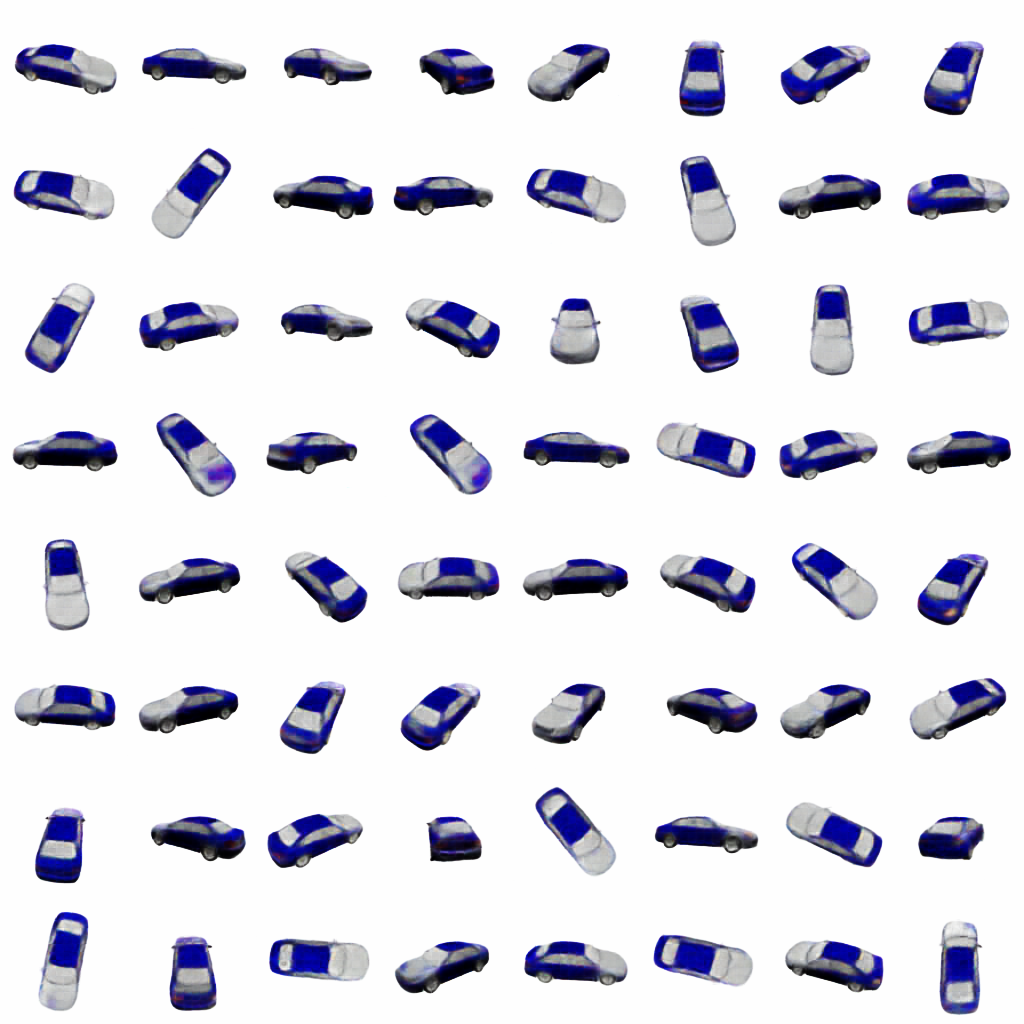}
\end{subfigure}
   \hfill
\begin{subfigure}{.32\linewidth}
    \centering
    \includegraphics[width=0.99\textwidth]{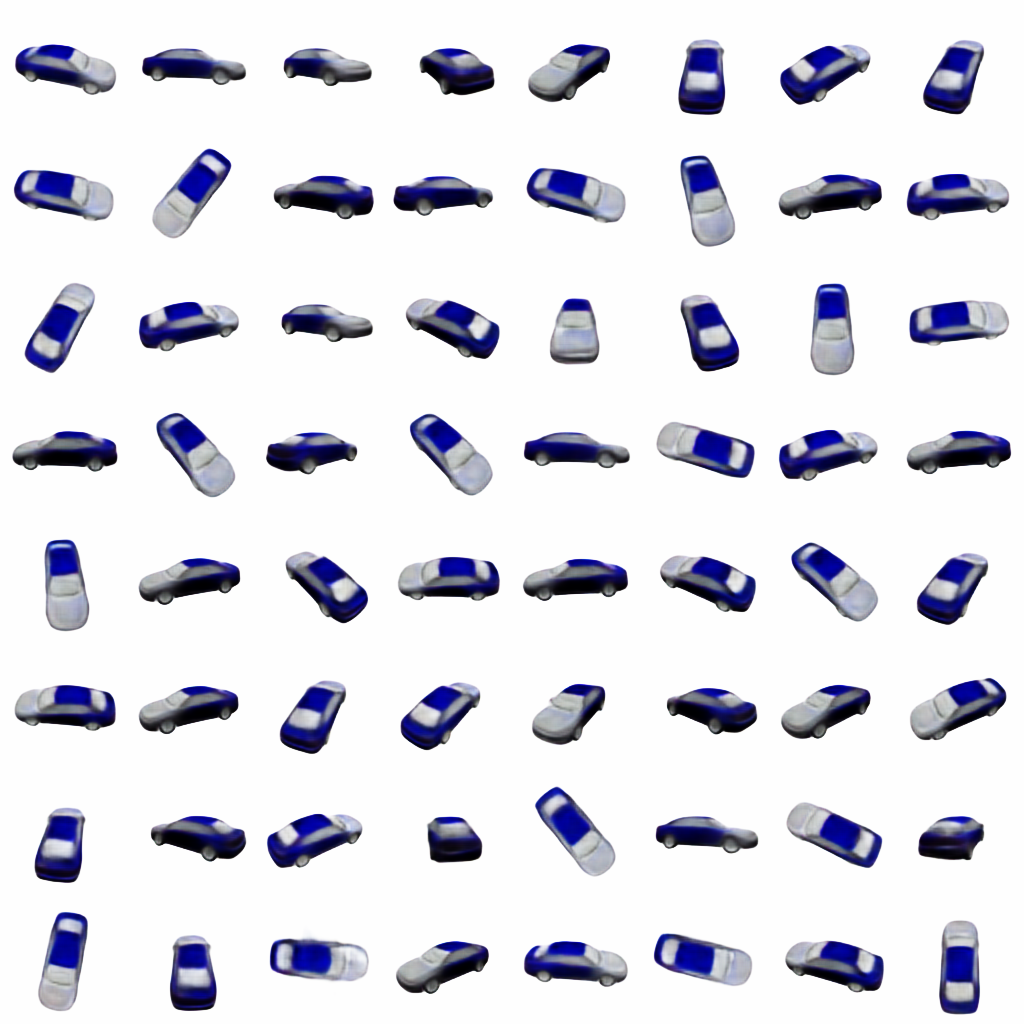}
\end{subfigure}

\begin{subfigure}{.32\linewidth}
    \centering
    \includegraphics[width=0.99\textwidth]{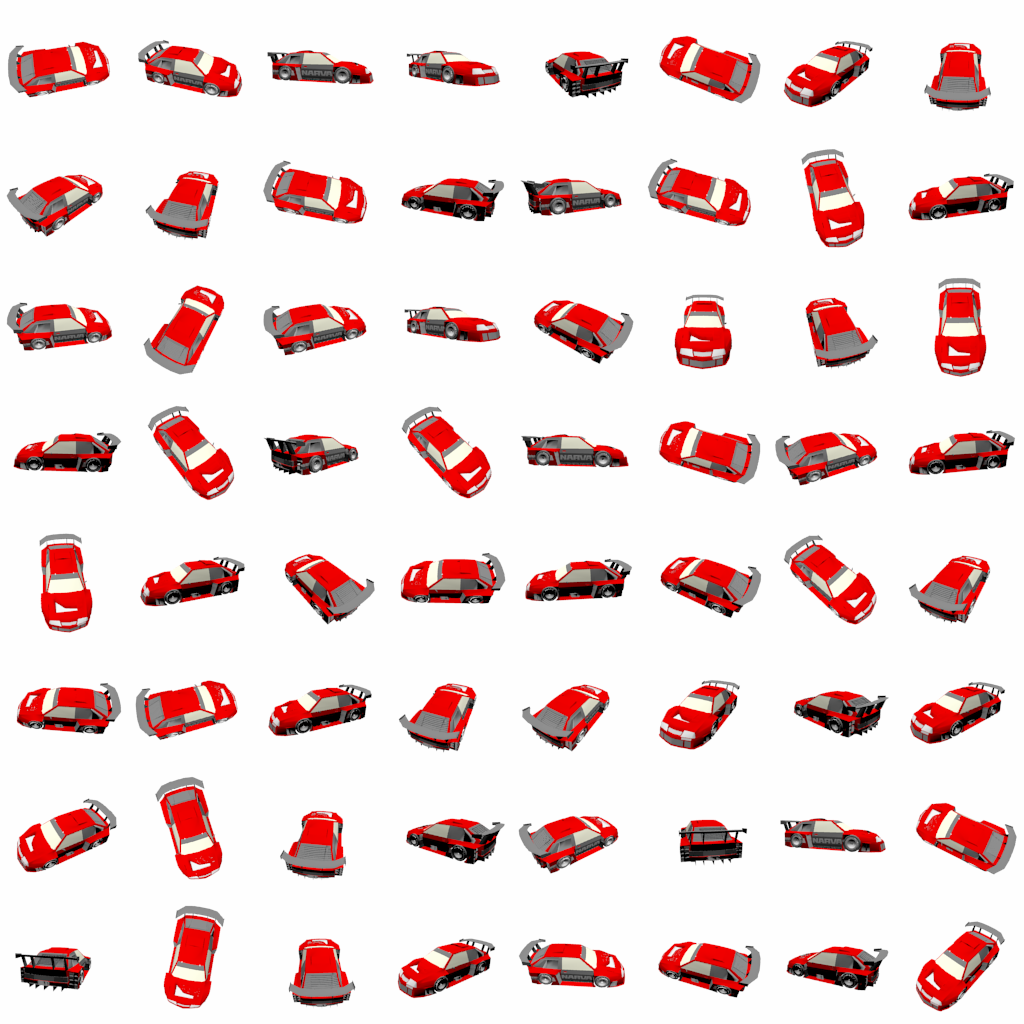}
    \caption{\small Ground Truth}\label{fig:image31}
\end{subfigure}
    \hfill
\begin{subfigure}{.32\linewidth}
    \centering
    \includegraphics[width=0.99\textwidth]{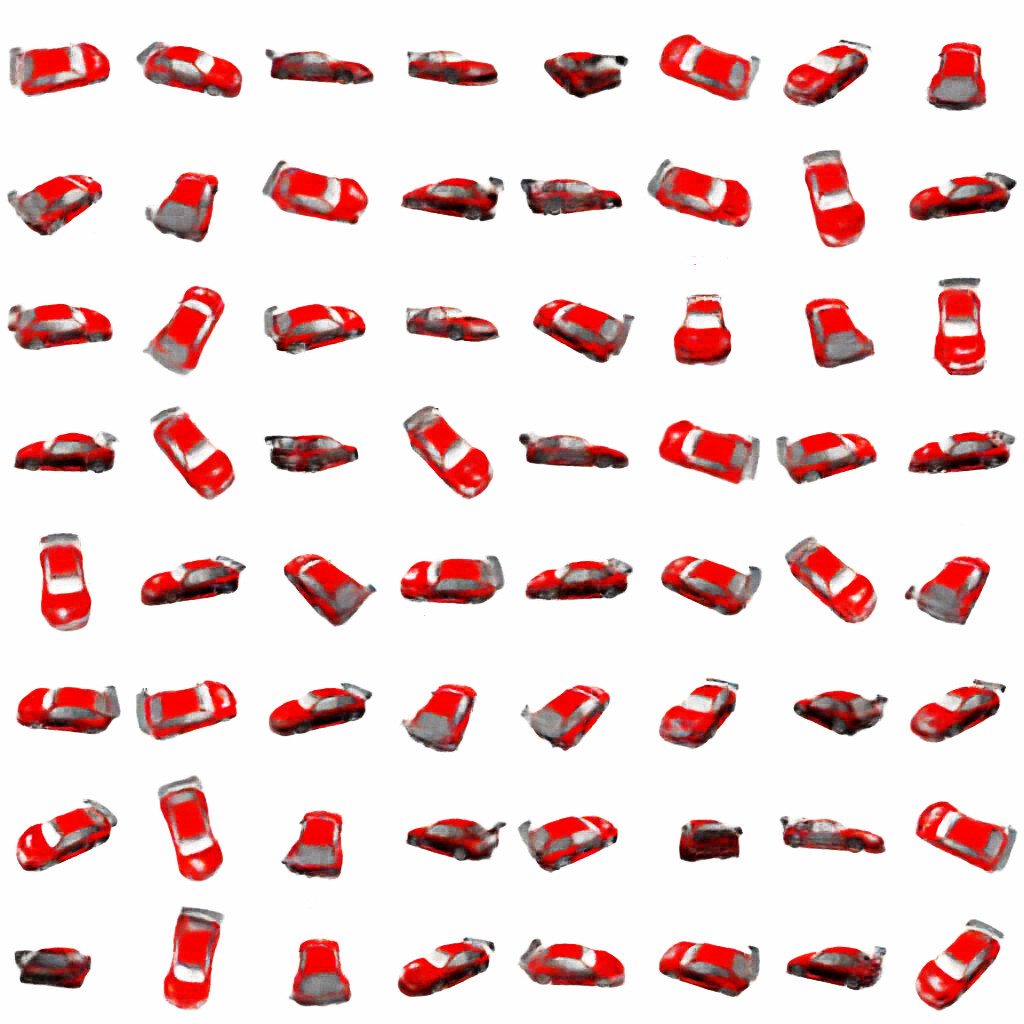}
    \caption{\small GQN}\label{fig:image32}
\end{subfigure}
   \hfill
\begin{subfigure}{.32\linewidth}
    \centering
    \includegraphics[width=0.99\textwidth]{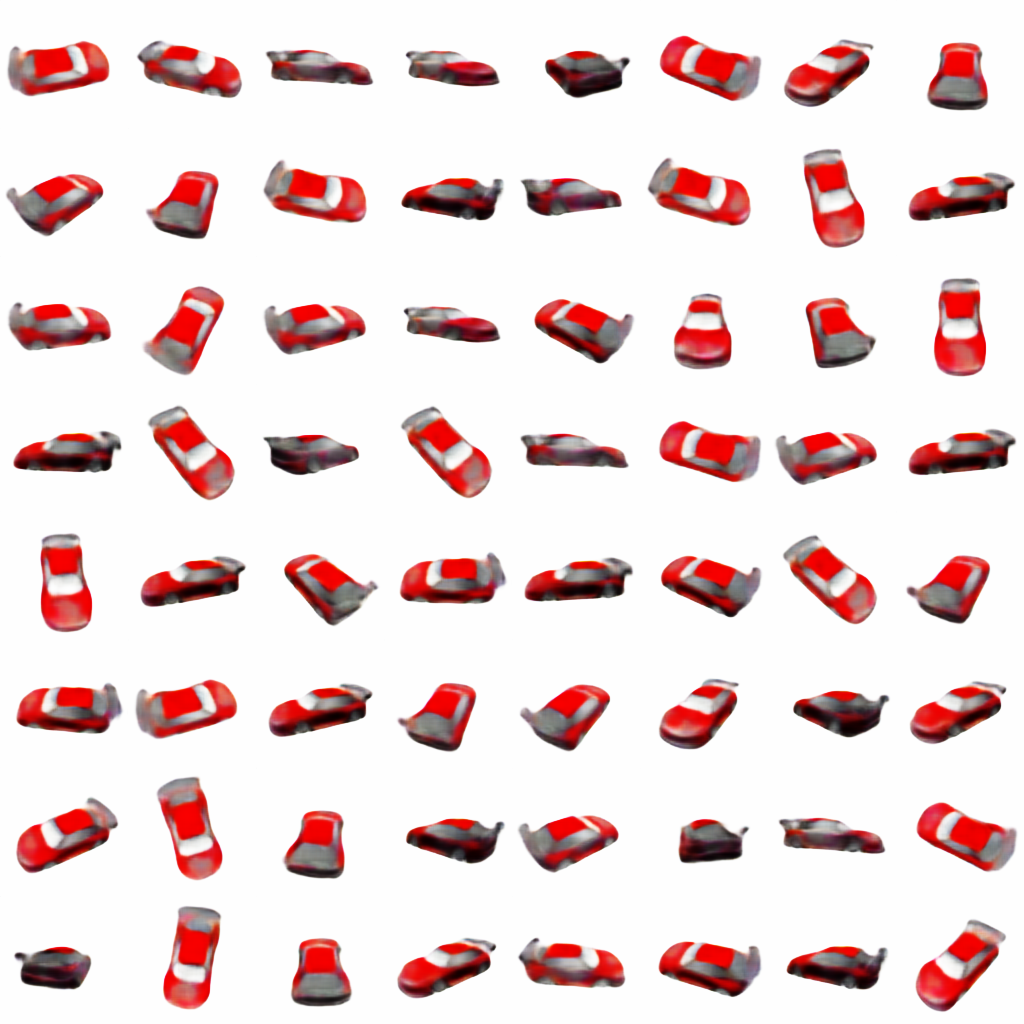}
    \caption{\small Ours}\label{fig:image33}
\end{subfigure}
\caption{\small Additional novel view synthesis results on ShapeNet v2 car dataset.}
\label{fig:car}
\end{figure*}

\begin{figure*}[ht!]
\centering
\begin{subfigure}{.32\linewidth}
    \centering
    \includegraphics[width=0.99\textwidth]{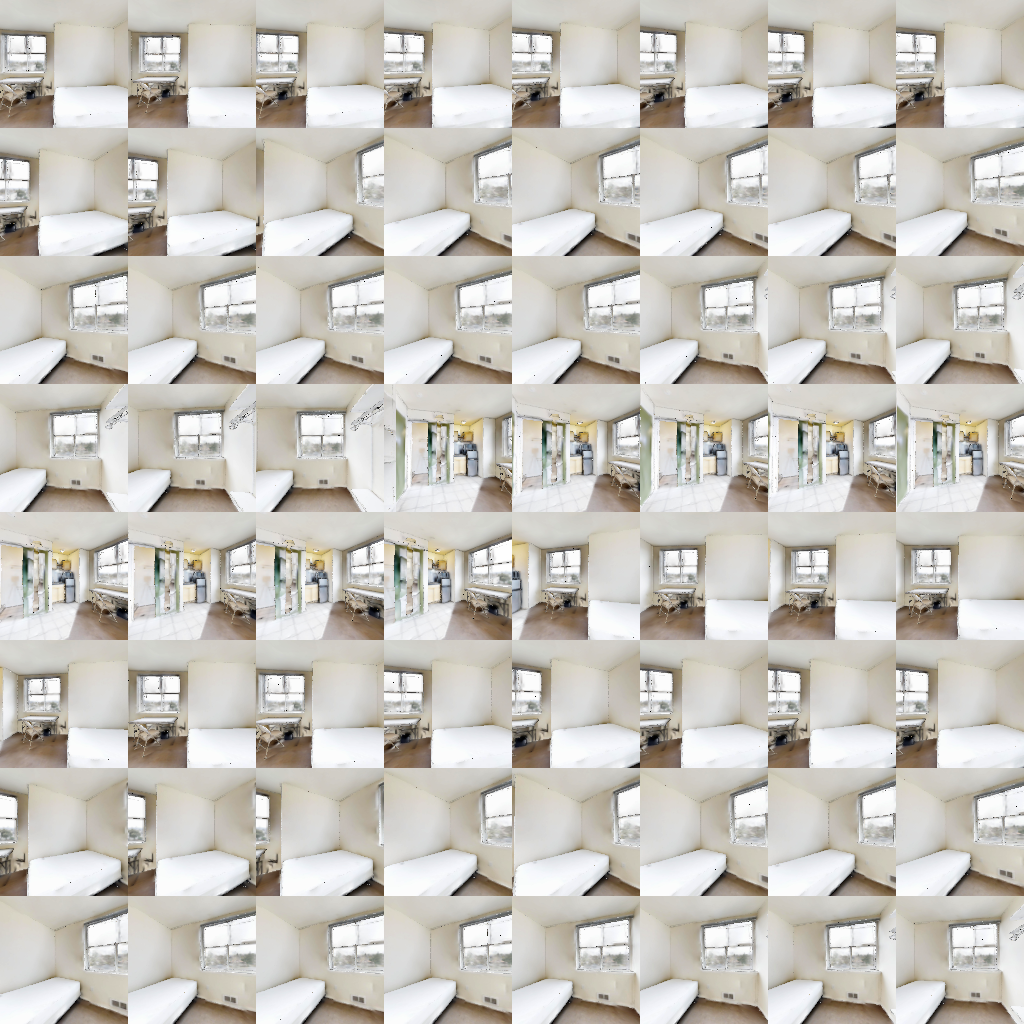}
\end{subfigure}
    \hfill
\begin{subfigure}{.32\linewidth}
    \centering
    \includegraphics[width=0.99\textwidth]{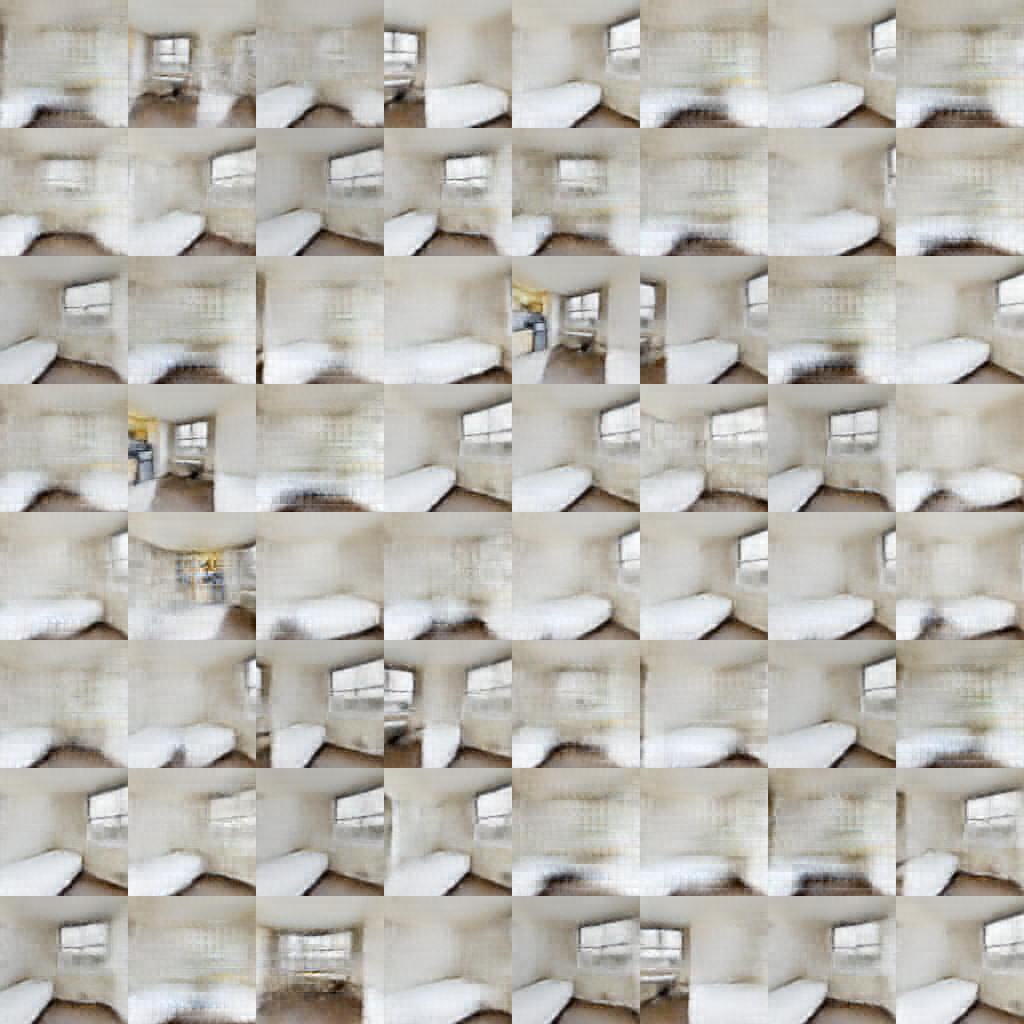}
\end{subfigure}
   \hfill
\begin{subfigure}{.32\linewidth}
    \centering
    \includegraphics[width=0.99\textwidth]{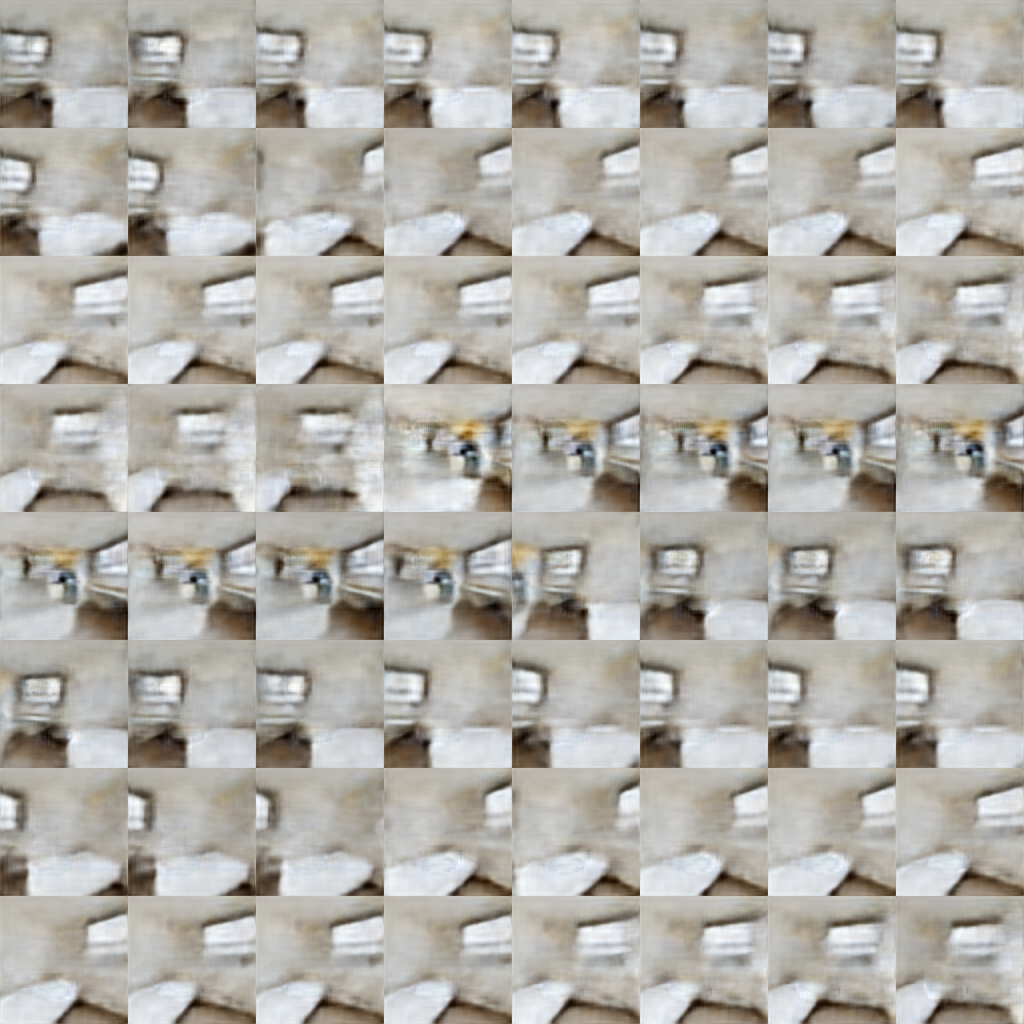}
\end{subfigure}

\begin{subfigure}{.32\linewidth}
    \centering
    \includegraphics[width=0.99\textwidth]{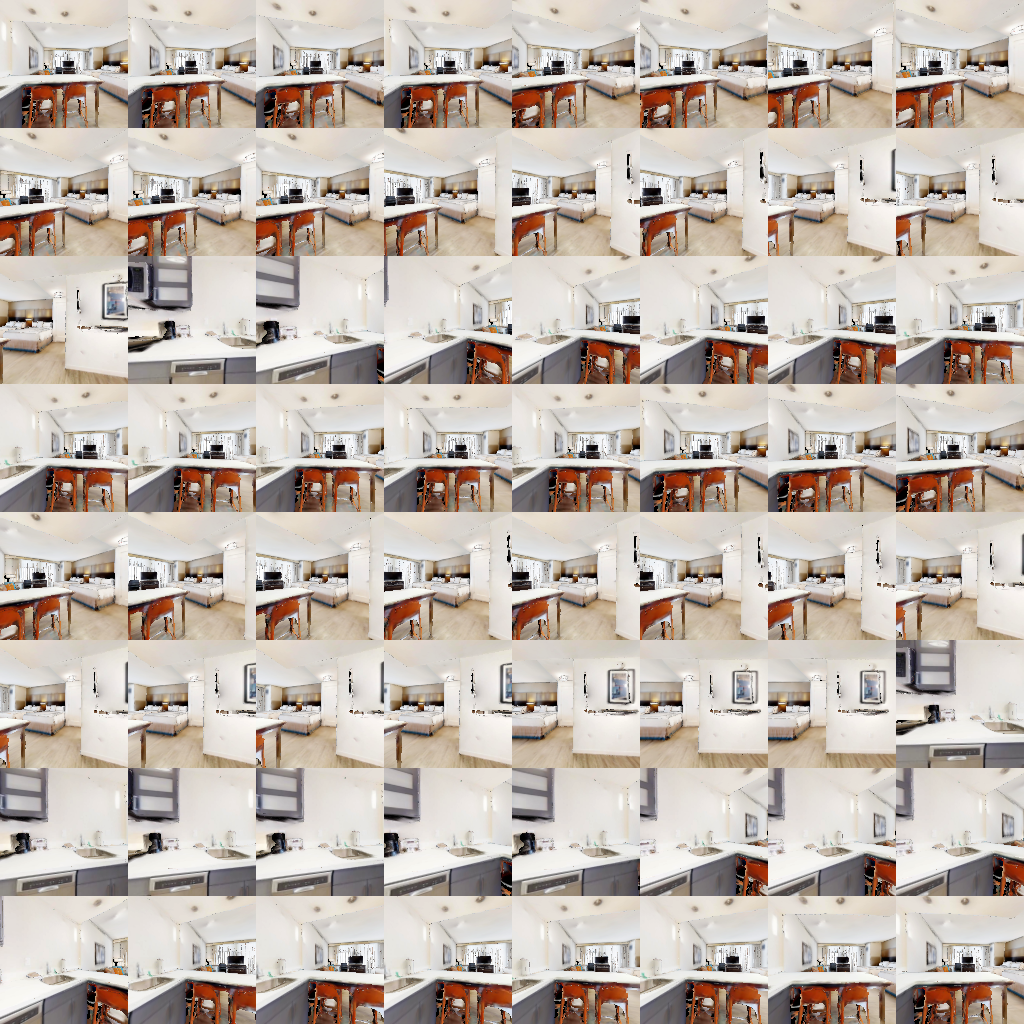}
\end{subfigure}
    \hfill
\begin{subfigure}{.32\linewidth}
    \centering
    \includegraphics[width=0.99\textwidth]{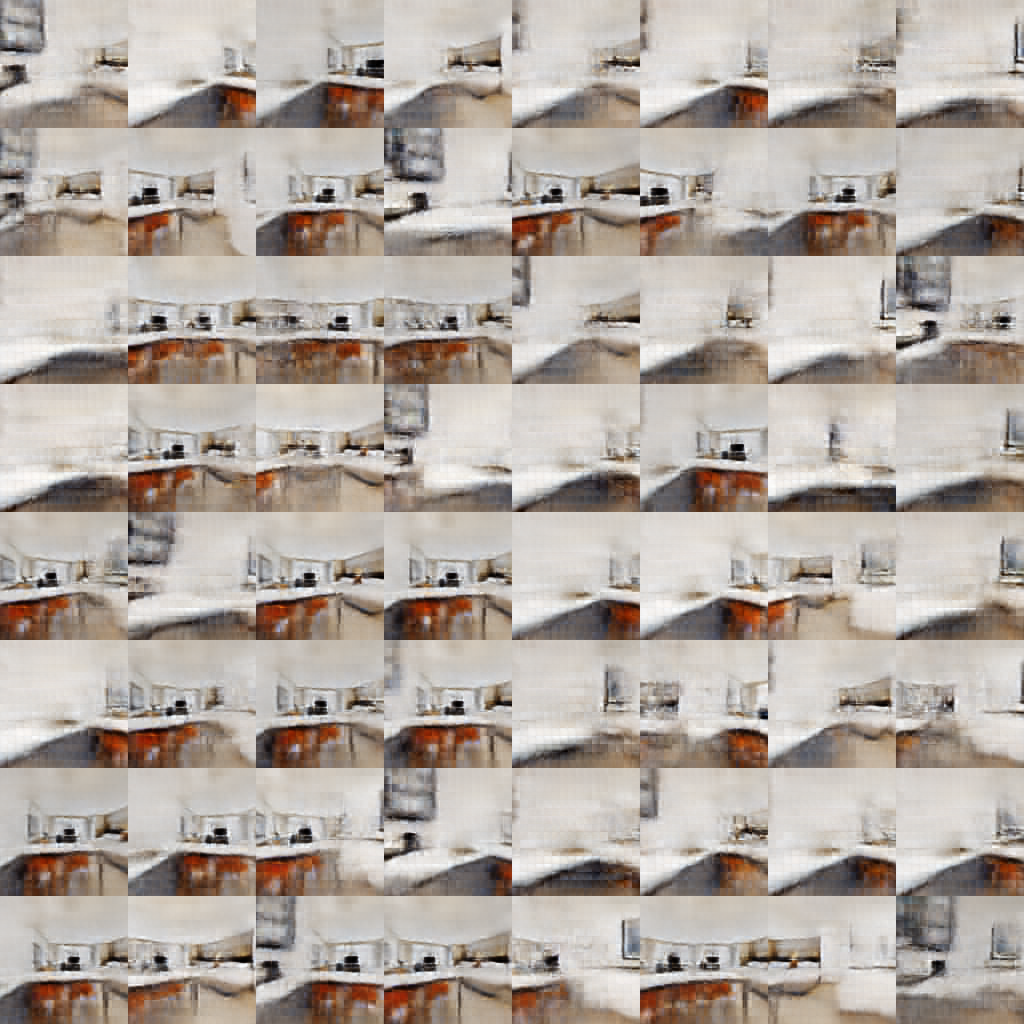}
\end{subfigure}
   \hfill
\begin{subfigure}{.32\linewidth}
    \centering
    \includegraphics[width=0.99\textwidth]{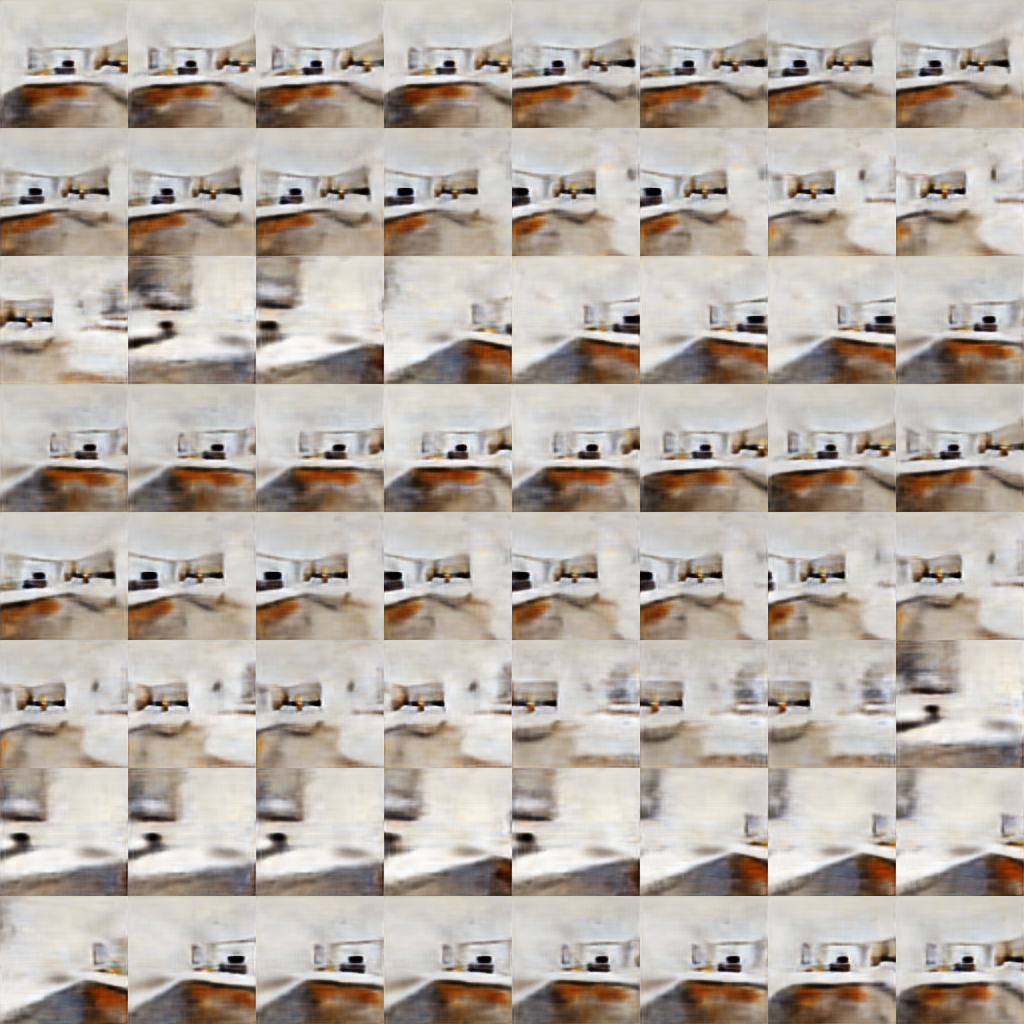}
\end{subfigure}

\begin{subfigure}{.32\linewidth}
    \centering
    \includegraphics[width=0.99\textwidth]{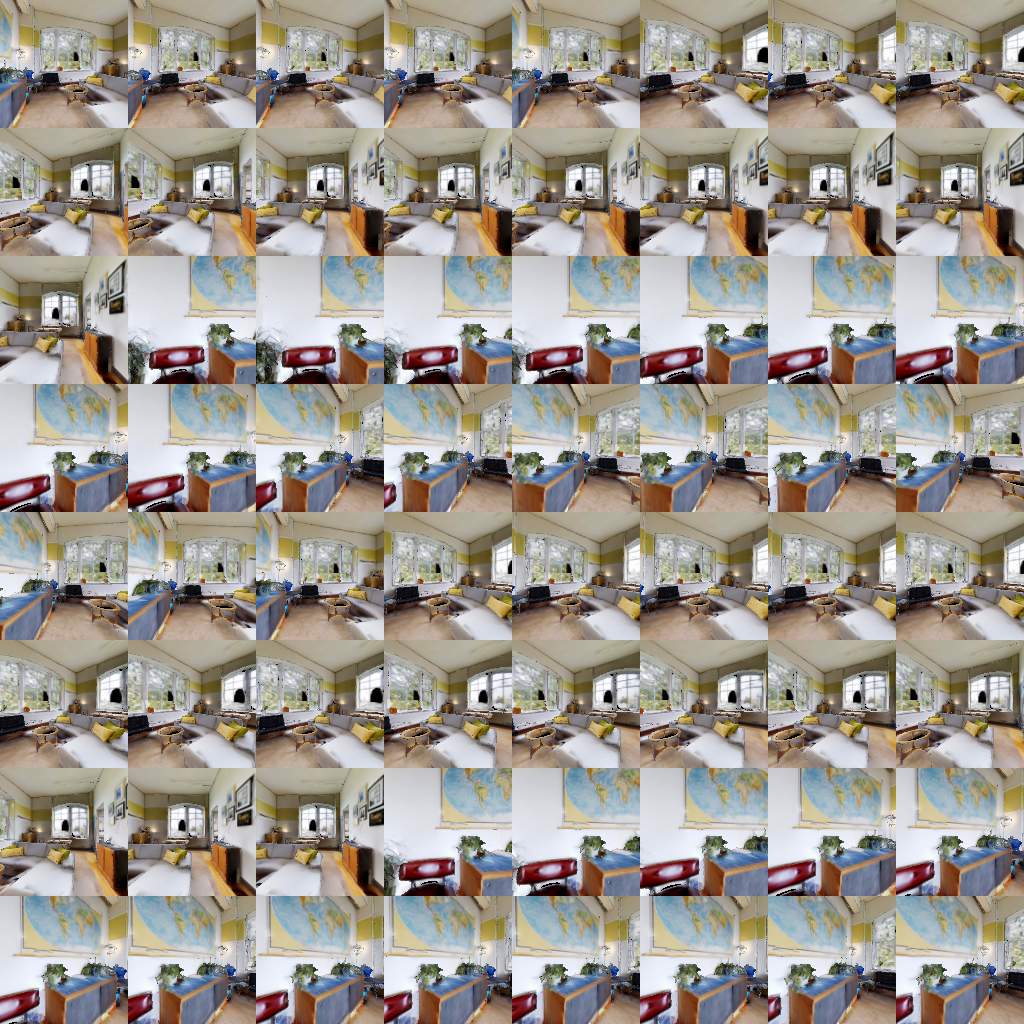}
    \caption{\small Ground Truth}\label{fig:image31}
\end{subfigure}
    \hfill
\begin{subfigure}{.32\linewidth}
    \centering
    \includegraphics[width=0.99\textwidth]{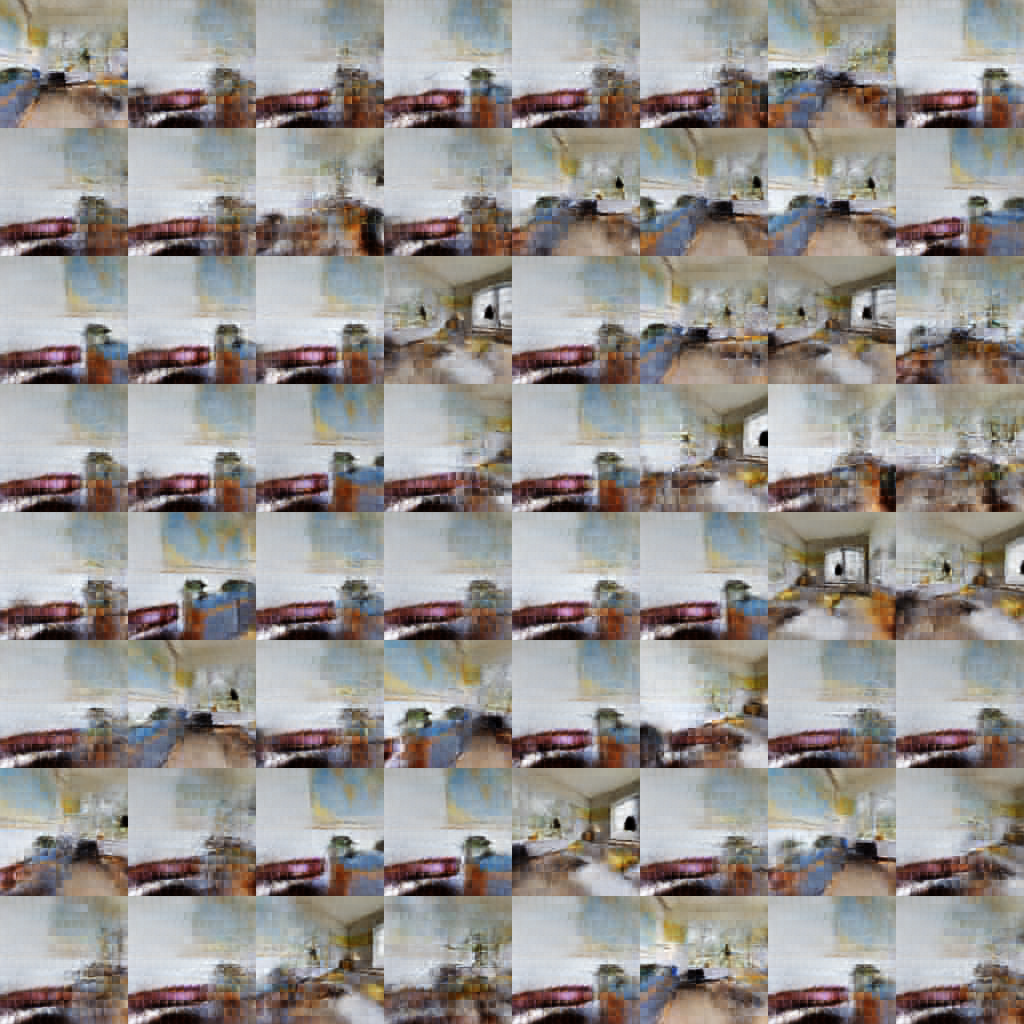}
    \caption{\small GQN}\label{fig:image32}
\end{subfigure}
   \hfill
\begin{subfigure}{.32\linewidth}
    \centering
    \includegraphics[width=0.99\textwidth]{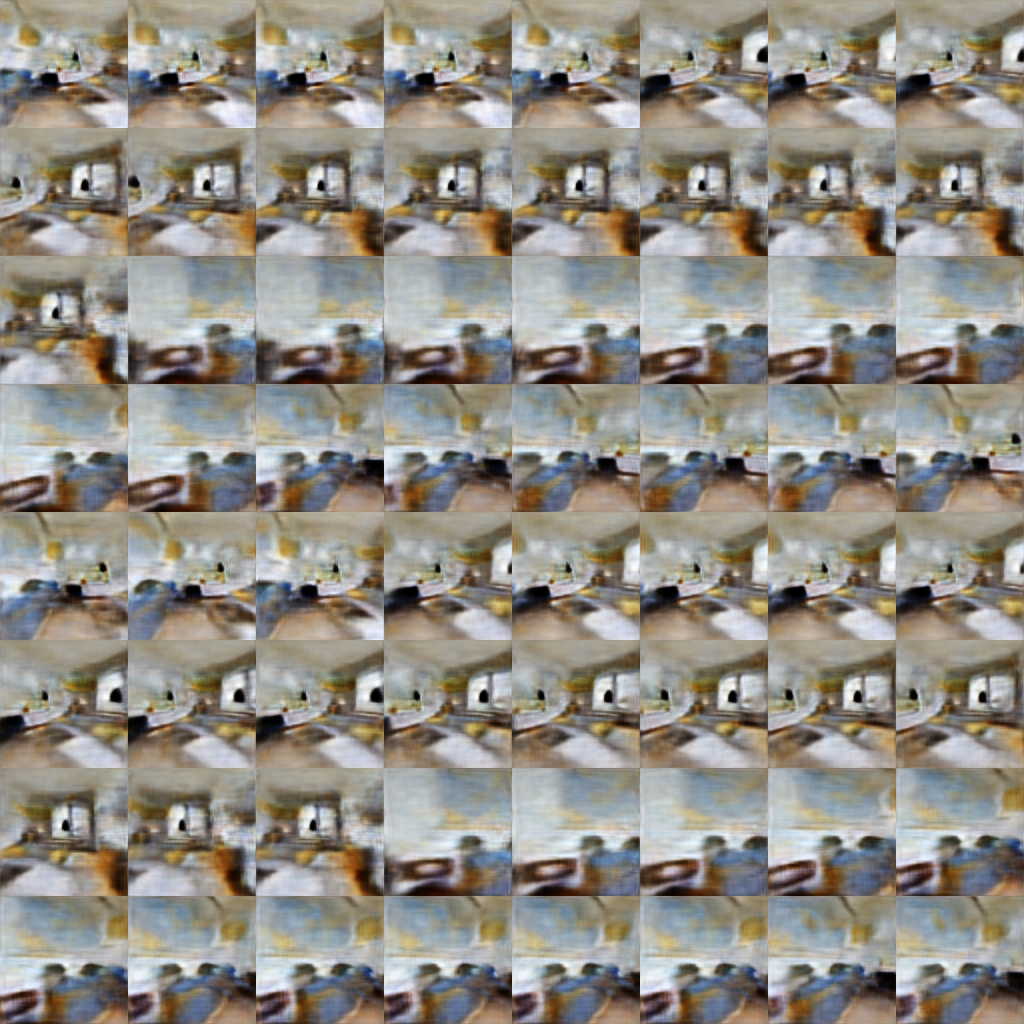}
    \caption{\small Ours}\label{fig:image33}
\end{subfigure}
\caption{\small Additional novel view synthesis results on Gibson rooms dataset.}
\label{fig:realroom}
\end{figure*}

We show more novel view synthesis results for GQN rooms, ShapeNet car and Gibson rooms in Figures \ref{fig:room}, \ref{fig:car}, \ref{fig:realroom}.

\subsection{Reconstructed image under different noise magnitude}
\begin{figure*}[ht!]
\centering
\includegraphics[width=0.9\textwidth]{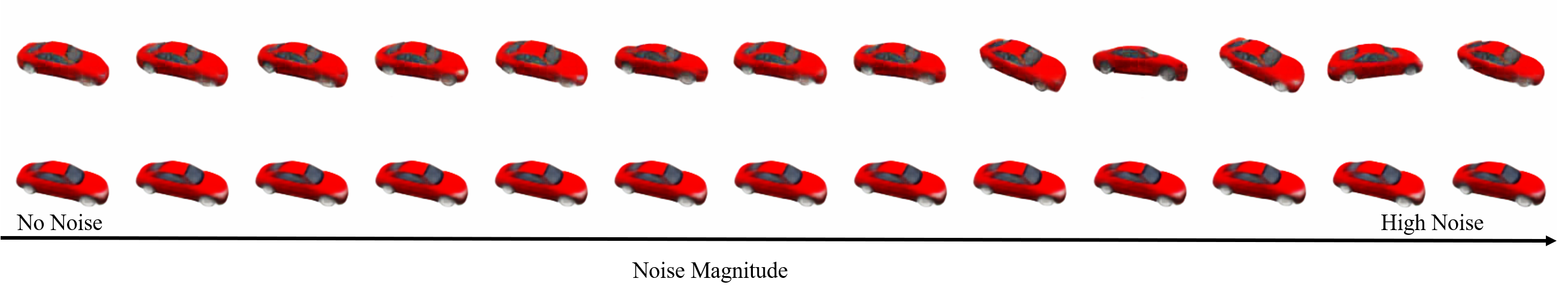}
\caption{\small Reconstructed image at different noise levels. Top: Results from GQN (which uses low dimensional $\sin(\alpha)$, $\cos(\alpha)$ to represent angles); Bottom: Our results with learned camera pose representation. From left to right: Gradually increasing the noise added to the camera pose representation.}
\label{fig:car_noise}
\end{figure*}

In Figure \ref{fig:car_noise}, we show the reconstructed images at different noise levels using our model with learned camera pose representation and GQN (which uses predefined low dimensional sinusoidal function to represent rotation). We can see that our model can reconstruct image with correct pose even with high noise while the poses in the reconstructed images of GQN model change a lot as noise increases. This agrees with our observations from the psnr curves and further prove that our learned camera pose representation is more robust to noise.

\subsection{Learning the camera pose representation by a fully connected neural network}
\begin{figure}[h!]
\centering
\includegraphics[width=0.45\textwidth]{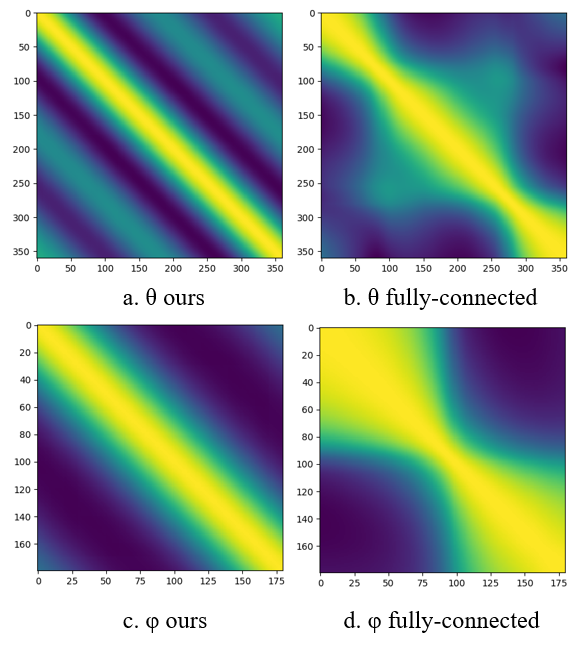}
\caption{\small Inner product of the learned camera pose representation. Each figure shows the inner product matrix between the camera pose representation at different positions (check \cite{wang2021position} for more details). {\em Left}: our camera pose representation; {\em Right}: Camera pose representation from a fully connected neural network. {\em Top}: Embedding for angle $\theta$; {\em Bottom}: Embedding for angle $\phi$.}
\label{fig:theta_phi}
\end{figure}

\begin{figure}[h!]
\centering
\includegraphics[width=0.45\textwidth]{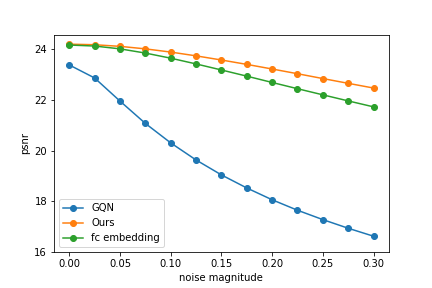}
\caption{\small PSNR over magnitude of noise added to representations, learned from ShapeNet car dataset.}
\label{fig:car_supp}
\end{figure}

As a comparison, we replace our proposed camera pose representation by a fully connected neural network on ShapeNet car dataset. Specifically, we encode each angle by a 2-layer fully-connected neural network. The first layer has a length of 128 the second layer has a length of 96 (which is same to our embedding). We use leaky relu as the activation function. As shown in Figure \ref{fig:theta_phi}, this embedding is also a high-dimensional one but it doesn't has the translation invariance \cite{wang2021position} as in our learned representation. Figure \ref{fig:car_supp} shows the PSNR over the magnitude of noise added to representations. The representation using a fully connected neural network works better than the plain low dimension embedding used in GQN in terms of robustness to noise. But it still performs worse than our design, which is regulated by the rotation loss. As for the camera pose estimation, using the representation from a fully connected neural network gives a testing error of \ang{3.63}, which is lower than the results of all the other hand designed representations but still higher than the result of our design (with testing error \ang{2.85}). The results show that learning a high-dimenstional representation is better than the low-dimensional hand designed ones and enforcing the translation invariance using rotation loss can further improve the results.  

\end{document}